\documentclass[lettersize,journal]{IEEEtran}
\usepackage{amsmath,amsfonts}
\usepackage{algorithmic}
\usepackage{algorithm}
\usepackage{array}
\usepackage[caption=false,font=normalsize,labelfont=sf,textfont=sf]{subfig}
\usepackage{textcomp}
\usepackage{stfloats}
\usepackage{url}
\usepackage{verbatim}
\usepackage{graphicx}
\usepackage{cite}
\usepackage[dvipsnames,table]{xcolor}
\usepackage[breaklinks,colorlinks]{hyperref}

\usepackage{tikz}
\usepackage{scalerel}
\usepackage{makecell}
\usepackage[nobiblatex]{xurl}

\usepackage{booktabs}
\usepackage[capitalize]{cleveref}

\definecolor{gold}{HTML}{FBF2D2}
\definecolor{silver}{HTML}{DDDDDD}
\definecolor{bronze}{HTML}{EED2B8}

\definecolor{goldD}{HTML}{D9AE13}
\definecolor{silverD}{HTML}{909090}
\definecolor{bronzeD}{HTML}{9A5F26}

\newcommand{\medal}[3]{\tikz[baseline=(char.base)]{\node[rounded corners=1pt,fill=#1,draw=#2,inner sep=1pt](char){#3};}}

\NewDocumentCommand\switch{}{\scaleto[10pt]{\includegraphics{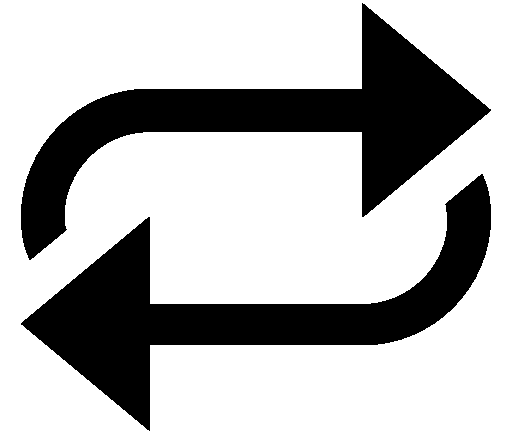}}{8pt}}

\newcommand{\bm}[2]{
    \ifcase#1\or
      {\medal{gold}{goldD}{\textbf{#2}}}
    \or 
      {\medal{silver}{silverD}{#2}}
    \or 
      {\medal{bronze}{bronzeD}{#2}}
    \else 
      #2
    \fi\ignorespaces
}

\definecolor{falseNeg}{HTML}{3d9af5}
\definecolor{falsePos}{HTML}{fa5757}

\newcommand{\meanwithstd}[2]{\specialcell[c]{#1 \\[-1pt] {\footnotesize ($\pm$ #2)}}}

\newcommand{\specialcell}[2][c]{%
  \begin{tabular}[#1]{@{}c@{}}#2\end{tabular}}

\definecolor{opti}{HTML}{4fffaa}

\newcommand{\resInc}[2]{\specialcell[c]{#1 \\[-2pt] \textcolor[rgb]{0,0.75,0.25}{{+#2}}}}
\newcommand{\resDec}[2]{\specialcell[c]{#1 \\[-2pt] \textcolor[rgb]{1,0,0}{{-#2}}}}
\newcommand{\resSame}[1]{\specialcell[c]{#1 \\[-2pt] \textcolor{sameRes}{0}}}

\newcommand{\inc}[1]{\textcolor[rgb]{0,0.75,0.25}{{+#1}}}

\definecolor{sameRes}{HTML}{5C5B59}

\definecolor{backbone}{HTML}{1b7ee0}
\definecolor{train}{HTML}{021975}

\let\titleold\title
\renewcommand{\title}[1]{\titleold{#1}\newcommand{\thetitle}{#1}}
\def\maketitlesupplementary
   {
   \newpage
       \twocolumn[
        \centering
        \Large
        \textbf{\thetitle}\\
        \vspace{0.5em}Supplementary Material \\
        \vspace{1.0em}
       ] 
   }

\definecolor{confBlue}{HTML}{1090EB}

\newcommand{\change}[1]{#1}

\begin{document}

\title{Be the Change You Want to See: Revisiting Remote Sensing Change Detection Practices}


\author{Blaž Rolih, Matic Fučka, Filip Wolf, Luka Čehovin Zajc
\thanks{(Corresponding author: Blaž Rolih)}
\thanks{The authors are with the University of Ljubljana, Faculty of Computer and Information Science, 1000 Ljubljana, Slovenia
(e-mail: blaz.rolih@fri.uni-lj.si)}
}

\markboth{Preprint}%
{Rolih \MakeLowercase{\textit{et al.}}: Be the Change You Want to See}


\maketitle

\begin{abstract}
Remote sensing change detection aims to localize semantic changes between images of the same location captured at different times. In the past few years, newer methods have attributed enhanced performance to the additions of new and complex components to existing architectures. Most fail to measure the performance contribution of fundamental design choices such as backbone selection, pre-training strategies, and training configurations. We claim that such fundamental design choices often improve performance even more significantly than the addition of new architectural components. Due to that, we systematically revisit the design space of change detection models and analyse the full potential of a well-optimised baseline. We identify a set of fundamental design choices that benefit both new and existing architectures. Leveraging this insight, we demonstrate that when carefully designed, even an architecturally simple model can match or surpass state-of-the-art performance on six challenging change detection datasets. Our best practices generalise beyond our architecture and also offer performance improvements when applied to related methods, indicating that the space of fundamental design choices has been underexplored. Our guidelines and architecture provide a strong foundation for future methods, emphasizing that optimizing core components is just as important as architectural novelty in advancing change detection performance. Code: \change{\url{https://github.com/blaz-r/BTC-change-detection}}.

\end{abstract}


\section{Introduction}
\label{sec:intro}

\IEEEPARstart{R}{emote} sensing change detection aims to detect and localize semantic changes between images of the same region. It is one of the fundamental tasks in remote sensing, providing insights into natural and human-driven processes such as deforestation, urban expansion, and natural disasters~\cite{chen2021bit, daudt2018urban}, which have a significant impact on climate, ecosystems, and human life~\cite{zheng2024segAnyCh, zhu2022rsLandChange, hansch2024eo4climate}. As remote sensing data availability grows, so does the need for efficient and accurate change detection methods.

\begin{figure}[t]
    \centering
    \includegraphics[width=1\linewidth]{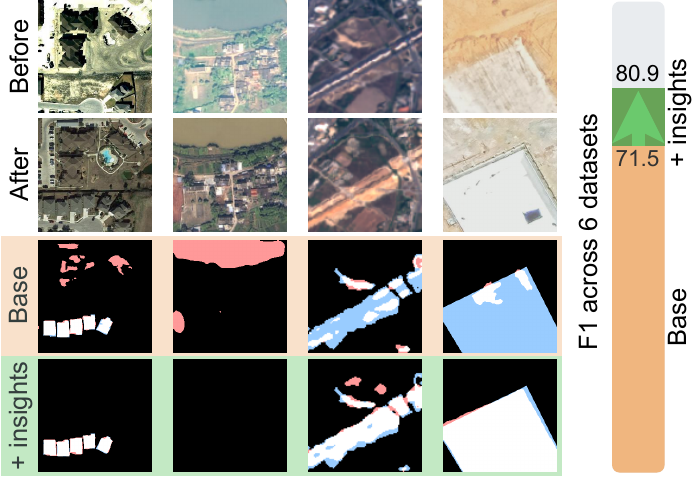}
    \caption{Comparison of change detection performance. The first two rows contain the input image pair, the third row depicts predictions of the base model, and the last row depicts predictions of the optimised model, incorporating guidelines from our analysis. False positives \textcolor{falsePos}{(red)} and false negatives \textcolor{falseNeg}{(blue)} highlight differences in prediction quality. The F1 score is reported across six different datasets. The improvement of $9.4~\%$ demonstrates how these fundamental yet often overlooked design choices greatly enhance performance (More information in~\Cref{sub:together}).}
    \label{fig:title}
\end{figure}

Change detection can be approached in various ways, but the dominant paradigm in remote sensing focuses on the bi-temporal setting in which a pair of images is analysed. Recently proposed deep-learning-based change detection methods~\cite{zhang2024bifa, bandara2025ddpmcd, bandara2022changeFormer, chen2021bit, feng2023dminet, li2023a2net} largely attribute performance gains to complex architectural innovations and fail to measure the performance contributions of fundamental design choices like backbone selection and training configurations. We hypothesize that a significant part of the observed performance improvements comes from making the correct fundamental design choices rather than architectural innovations. 
On top of that, many methods nowadays still underutilise pre-training, employ inconsistent data augmentations, and rarely experiment with alternative training strategies despite recent advancements~\cite{liu2021swin,liu2022swinv2,liu2022convnet}, meaning there is an excellent opportunity for improvement in picking the right design choices.
While some studies have also explored new backbone and training choices~\cite{chen2021bit, zheng2021changeStar, zhang2022swinsunet, chen2024changeMamba}, they often do so accompanied by larger architectural changes, making it difficult to quantify their standalone contributions. We hypothesize that carefully selecting and optimizing these elements significantly impacts change detection performance.

In our work, we build a model from scratch (starting from a simple baseline), varying only the base design choices. We iteratively refine the simple base model, independently examining each design choice's impact on the overall performance. More specifically, we focus on backbone architecture, backbone size, pre-training, data augmentation techniques, loss functions, and learning rate schedulers. While many of these techniques were studied in other domains~\cite{heckler2023backbones, orsic2019in1ksemSeg, zhang2023dino, rolih2024supersimplenet}, such as pre-training on natural images\footnote{The term \textit{natural images} typically refers to images captured at ground-level and containing diverse real-world scenes, such as those found in the ImageNet dataset. We use this term to distinguish these images from the top-down view of satellite and aerial images.}, they have not yet been comprehensively evaluated in change detection. This process leads to an optimized yet architecturally simple change detection model, which we refer to as \textit{BTC} (\textbf{B}e \textbf{T}he \textbf{C}hange). After applying all gathered insights, the performance of \textit{BTC} compared to the baseline model across six datasets increases by $9.4$ percentage points (p.p.), as illustrated in Figure~\ref{fig:title}. Furthermore, we show that applying our guidelines to existing remote sensing foundation models and change detection-specific architectures improves their performance.

We conduct extensive experiments on six standard change detection datasets to validate our insights and demonstrate their broad applicability. Our results show that a simple but well-optimised \textit{BTC} architecture achieves performance comparable to, or even surpassing, state-of-the-art methods, particularly on smaller datasets. This reinforces our claim that optimal fundamental design choices have been underutilised and that recent performance gains cannot be solely attributed to architectural novelty. The lack of awareness about the impact of these core components, coupled with the absence of systematic analyses, has likely contributed to this oversight. 
The contributions of this work are as follows:
\begin{itemize}
    \item We \textbf{systematically analyse} key design choices of base components in change detection methods. This includes various backbone aspects, pre-training, data augmentations, loss functions, and learning rate schedulers. 
    Our findings highlight the overlooked impact of these fundamental components and help us \textbf{establish best practices} for their use in the future.
    \item Our study goes beyond standard optimisation by showing \textbf{generalisation to related methods}. By applying our insights to remote sensing foundation models and other change detection methods, we show consistent improvements in their performance. This verifies that recent methods have underutilized these fundamental design principles and that architectural novelty alone does not account for observed improvements.
    \item We propose \textit{BTC}, an architecturally simple yet \textbf{powerful change detection model}. It leverages our observations to match or even outperform the current state-of-the-art. Due to its architectural simplicity, we believe it presents an excellent foundation for new architectures to build upon. 
\end{itemize}

\section{Related Work}
\label{sec:related_work}

\noindent \textbf{Change detection} has a long history, ranging as far as 50 years into the past~\cite{singh1989reviewCD}. Some of the early approaches used simple pixel-wise comparisons, statistical methods, or traditional machine learning~\cite{le2013urbanSar, metzger2023UCForecast}. The field has witnessed an increase in popularity with the introduction of deep learning methods. One of the first successful deep-learning-based approaches that outperformed traditional methods used convolutional architectures~\cite{daudt2018urban}. FC-Siam-Diff~\cite{daudt2018fcn} introduced a Siamese network as the encoder, with the idea that the latent representations of the images in a pair will be aligned if there are no semantic changes. This architecture, consisting of two encoders with shared weights, remains a widely used foundation in modern change detection.

With the introduction of new architectural elements~\cite{dosovitskiy2021vit, gu2023mamba}, contemporary change detection methods have advanced by integrating transformer models~\cite{chen2021bit, bandara2022changeFormer, zhang2022swinsunet} and other building blocks~\cite{liu2024vmamba, chen2024changeMamba, bandara2025ddpmcd}. Many, however, still rely on CNNs for feature extraction and use various attention approaches to improve performance~\cite{feng2023dminet, li2023a2net, chen2020levirStanet}. 
Given the challenge of fully training backbones from scratch on limited data~\cite{noman2024rsCDfromScratch}, the majority of such methods rely on pre-trained backbones, typically trained on ImageNet~\cite{deng2009imagenet}. BAN~\cite{li2024ban} uses a remote sensing foundation model instead of a regular backbone to enhance the features used in its change detection architecture. However, the impact of backbone architecture and its pre-training is rarely evaluated~\cite{zhang2022swinsunet}, despite its significance that we show in this paper.

\noindent \textbf{Pre-training} plays an important role in many remote sensing tasks~\cite{astruc2024anysat, astruc2024omnisat}, with most approaches leveraging self-supervised contrastive learning (CL)~\cite{ayush2021gessl, manas2021seco, mall2023caco} or masked image modelling (MIM) architectures~\cite{cong2022satmae, tang2024xscalemae}. The authors of GFM~\cite{mendieta2023gfm} additionally explored the combination of the MIM approach with the simultaneous distillation of features from an ImageNet pre-trained model.
However, these strategies do not always translate to optimal change detection performance~\cite{wang2024mtp, wang2022empiricalRSPt}.
While multitask pre-training (MTP)~\cite{wang2024mtp} has demonstrated strong results, its complexity and high computational cost can limit its practicality.
Moreover, most pre-training approaches focus solely on representation learning rather than optimizing the training process specifically for change detection.

Addressing this, some methods have introduced pre-training tailored to change detection~\cite{zhang2023ssl4cd, leenstra2021ssl_s2, mall2023caco, quan2023unified, zheng2023changen}. Authors of the Changen method~\cite{zheng2023changen} proposed the generation of synthetic changes to create a large-scale dataset for pre-training, achieving good results. Despite these efforts, many specialized pre-training techniques struggle to consistently outperform general-purpose backbones pre-trained on tasks like ImageNet classification~\cite{wang2022empiricalRSPt}.

\noindent \textbf{Systematic analysis and optimization of training strategies} was explored for convolutional networks~\cite{he2019cnnOpti, liu2022convnet}, large language models (LLMs)~\cite{yu2023llmOpti}, and, to a limited extent, remote sensing pre-training~\cite{wang2022empiricalRSPt, sosa2024effective}.
Corley et al.~\cite{corley2024cdReality} address some of the issues with current change detection methodologies and their evaluation protocols, though without an in-depth analysis of underlying factors.
Meanwhile, systematic research on architectural and training aspects for change detection remains sparse~\cite{wang2022empiricalRSPt, noman2024rsCDfromScratch}. Our paper thus contributes an extensive analysis with valuable insights into these challenges.

\section{Methodology}
\label{sec:method}

\noindent To clarify our evaluation approach, we first introduce the change detection framework we use in \Cref{subsec:cd-framework}. Next, we outline the evaluation protocol in \Cref{subsec:eval}, covering datasets and metrics. Finally, we provide key implementation details in \Cref{subsec:impl_det}.

\subsection{Change detection framework}
\label{subsec:cd-framework}

\begin{figure}
    \centering
    \includegraphics[width=1\linewidth]{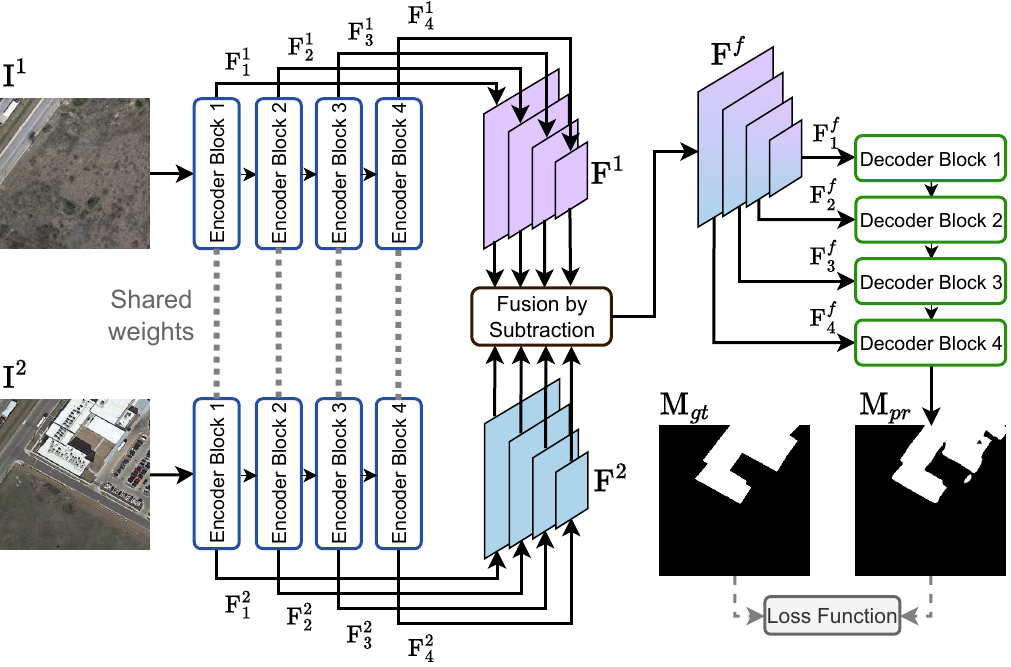}
    \caption{\change{Outline of our change detection method. Each image in the given pair ($\mathrm{I}^1, \mathrm{I}^2$) is processed using the same encoder. Then, the extracted features ($\mathrm{F}^1, \mathrm{F}^2$) are fused (in our case, by simple subtraction), and the resulting fused features $\mathrm{F}^f$ are decoded using the UPerNet decoder \cite{xiao2018upernet} to obtain the predicted change mask $\mathbf{M}_{pr}$.}}
    \label{fig:cd-arch}
\end{figure}

\noindent We use a simple architecture (illustrated in \Cref{fig:cd-arch}) to evaluate the impact of different design choices. The model consists of three main components: 1) The \textit{Siamese encoder} first extracts features from the pre-change and post-change images. 2) A \textit{Feature fusion module} then combines the information from these features into a single representation. 3) The \textit{Decoder} finally predicts the binary pixel-level change map based on the given representations.

A common choice for the \textit{encoder} (also called \textit{backbone}) are networks such as ResNet~\cite{he2016deep}, ViT~\cite{dosovitskiy2021vit}, and ConvNext~\cite{liu2022convnet}. Their key advantage is that they have pre-trained weights available from various pre-training tasks (e.g. segmentation, classification). We primarily use Swin~\cite{liu2021swin}, a choice experimentally validated later in the paper. The Siamese setup applies the same encoder with shared weights to both images, producing corresponding feature maps.

Rather than introducing additional fusion or decoder modifications, we focus on optimizing the encoder -- a crucial yet often overlooked aspect of change detection models. While many methods utilize standard backbones and refine fusion and decoder modules~\cite{zhang2022swinsunet, li2023a2net, bandara2022changeFormer, zhang2024bifa, chen2021bit}, our analysis demonstrates that backbone choice alone leads to substantial gains, making it a critical design consideration for future models. We preliminarily evaluated some non-complex fusion and decoder designs from prior work~\cite{daudt2018fcn, mendieta2023gfm, wang2024mtp} and adopted well-established and best-performing choices to ensure robust and fair comparison. 

The \textit{feature fusion module} is implemented as a simple element-wise \textit{subtraction} of features at the same hierarchical level: $\mathrm{F}^f_{i} = \mathrm{F}^{1}_i - \mathrm{F}^{2}_i$, where $\mathrm{F}^1$ and $\mathrm{F}^{2}$ are features from pre- and post-change images respectively, and $i$ represents different layer levels from the encoder.

For the \textit{decoder}, we use UPerNet~\cite{xiao2018upernet}, which efficiently processes multi-level fused features $\mathrm{F}^f$ to generate a change segmentation map. The change map is then thresholded at $0.5$ to produce the final binary change mask $\mathbf{M}_{pr}$.

\subsection{Evaluation regime}
\label{subsec:eval}

\noindent Each experimental setup is evaluated on six different change detection datasets (described in \Cref{subsec:datasets}). For consistency, all experiments use the same set of hyperparameters (provided in \Cref{subsec:impl_det}), and all model parameters are optimised (i.e. the backbone is never frozen).  

To ensure a fair comparison, each experiment is repeated with three different random seeds and the average F1 across these runs is reported. We also include \textit{standard deviation} in Supplementary S4. The evaluation is performed on the predefined test set of each dataset using the model from the final epoch.

\subsubsection{Datasets}
\label{subsec:datasets}

For a comprehensive and robust evaluation, we select six commonly used RGB change detection datasets from various sensors, covering diverse change types, imaging conditions, ground sample distances (GSD), and dataset sizes. \textit{SYSU}~\cite{shi2022sysuDSAMnet} is used as a candidate for changes of multiple types. Two datasets contain only building changes, the larger \textit{LEVIR}~\cite{chen2020levirStanet} and the smaller \textit{EGYBCD}~\cite{holail2023egybcd}. \textit{GVLM}~\cite{zhang2023gvlm} is used to evaluate landslide changes and \textit{CLCD}~\cite{li2022clcdMSCANET} is a smaller dataset focusing on cropland changes. Finally, \textit{OSCD}~\cite{daudt2018urban} is used to verify small-scale datasets with low-resolution Sentinel-2 imagery. 

\begin{itemize}
    \item \textit{SYSU} consists of 20,000\footnote{\label{p256}Number of $256 \times 256$ patches} pairs of image patches with a high resolution of $0.5~m$. It captures multiple change types, including buildings, groundwork, vegetation, and sea changes.
    \item \textit{LEVIR} contains over 10,000\footref{p256} image pair patches, captured over 5-14 years in 20 US regions, with a resolution of $0.5~m$. It is specifically designed for building change detection and includes only such labels.
    \item \textit{EGYBCD} is a smaller building change detection dataset with a little over 6,000\footref{p256} image pairs. It features images with a high resolution of $0.25~m$ captured in Egypt between 2015 and 2022.
    \item \textit{GVLM} is tailored for landslide change detection. It contains over 7,500\footref{p256} image pairs at a resolution of $0.59~m$ collected from 17 different regions worldwide.
    \item \textit{CLCD} targets cropland change detection. It includes approximately 2,400\footref{p256} images collected between 2017 and 2019. It has varying resolutions from $0.5~m$ to $2~m$ and focuses on visually challenging cropland changes.
    \item \textit{OSCD} is a low-resolution dataset with $10~m$ resolution images captured worldwide between 2015 and 2018 using the Sentinel-2 satellite. It consists of around 1,200\footnote{Number of $96 \times 96$ patches} image patch pairs, focusing only on urban changes.
\end{itemize}
More extensive details on these datasets can be found in Supplementary S1.

\subsubsection{Metrics}
\label{subsec:metrics}

Following standard practice~\cite{bandara2022changeFormer, wang2024mtp, chen2021bit, daudt2018fcn, zheng2023changen}, we evaluate change detection performance using the \textbf{binary F1} score, considering only \textbf{change class}. 
Some recent works deviate from this practice by treating the background as a separate class and computing the mean F1 score ($m\mathrm{F1}$).
Due to the inherent class imbalance, this inflates the F1 score and prevents fair comparison with other works.

For illustration, our best setup achieves an F1 of $80.9~\%$ in the binary case, while the $m\mathrm{F1}$ reaches $89.6~\%$. This discrepancy highlights why we urge researchers to adhere to the established protocol when implementing metrics.

A more in-depth discussion of the implementation and evaluation is present in Supplementary S2.

\subsubsection{Proposed evaluation protocol}
\label{subsec:protocol}

\change{
To promote consistency and reproducibility, we compile the above-stated details and outline the following recommended evaluation protocol for future change detection studies:}
\begin{itemize}
    \item \change{Use a diverse set of datasets covering different change types, resolutions, and acquisition settings;}
    \item \change{Ensure consistent training hyperparameters across experiments;}
    \item \change{Repeat each experiment with multiple random seeds and report the average performance;}
    \item \change{Evaluate models on a fixed test set;}
    \item \change{Use binary F1 score (focusing on the change class) to avoid performance inflation from class imbalance;}
    \item \change{Fully disclose the design decision, even the fundamental choices.}
\end{itemize}

\change{
Future methods can utilise our framework code (link is in the abstract), which already covers all the above-listed aspects.
}

\subsection{Baseline training details}
\label{subsec:impl_det}

\noindent For our baseline, we start with Swin-T~\cite{liu2021swin} as the \textit{encoder (backbone)}. Consistent with related work~\cite{chen2021bit, zhang2022swinsunet, zheng2021changeStar, chen2024changeMamba, chen2020levirStanet}, the backbone is initialized with \textit{ImageNet1k}~\cite{deng2009imagenet} pre-trained weights, unless stated otherwise. All other model weights follow the standard random initialisation. Input images are cropped to $256 \times 256$ pixels, except for OSCD, where, following the original protocol~\cite{daudt2018urban, daudt2018fcn}, images are cropped to $96\times96$ pixels and then rescaled to $256 \times 256$ pixels. The model is trained for 100 epochs with a batch size of 32 using the AdamW~\cite{Loshchilov2017adamw} optimiser with a base learning rate of $10^{-4}$ and with a weight decay of $10^{-4}$. The only exception is OSCD, where we only train for 50 epochs due to a small training set. Following the literature~\cite{wang2024mtp, chen2021bit, bandara2022changeFormer, zheng2023changen}, the model is trained in a supervised fashion using binary \textit{cross-entropy} loss, \textit{with no scheduler} and \textit{no augmentations}, unless stated otherwise. These hyperparameters remain constant throughout all experiments to ensure robust comparison.

\section{Backbone Design Choice Analysis}
\label{sec:bbone_opt}

\noindent The first set of design choices concerns the backbone, a cornerstone of many change detection methods that is often underutilised. We analyse the choice of pre-training, architecture, and model size in the following subsections.

\subsection{Pre-training dataset}
\label{subsec:pt-data}

\noindent ImageNet1k~\cite{deng2009imagenet} pre-trained encoders are very common in related work~\cite{chen2021bit, zhang2022swinsunet, zheng2021changeStar, chen2024changeMamba, chen2020levirStanet}, but whether they are the best-suited for change detection remains unclear. Swin offers multiple pre-training options, allowing us to assess the impact of dataset and task selection. We compare ADE20k~\cite{zhou2017ade20k} and Cityscapes~\cite{cordts2016cityscapes} (semantic segmentation), COCO~\cite{lin2014coco} (panoptic segmentation), ImageNet1k~\cite{deng2009imagenet} - IN1k - (classification) and EuroSat~\cite{helber2019eurosat,helber2018introducingEurosat} (RS classification). Only the best-performing task-dataset pairs are reported, while complete results and implementation details are in Supplementary S3-A. Remote sensing foundation models are analysed separately in \Cref{sec:transfer}.

\Cref{tab:dataset} shows that segmentation pre-training outperforms classification, with Cityscapes-based semantic segmentation yielding the best results. Interestingly, EuroSat pre-training does not surpass ImageNet1k pre-training. Semantic segmentation pre-training likely enhances performance because change detection is a segmentation task. These findings suggest that pre-training for a similar task is more important than pre-training on domain data, which is an important insight that could guide future remote sensing pre-training schemes.

\begin{table}[!h]
\caption{Comparison of different pre-training datasets for the Swin backbone. \textit{None} means that no pre-training dataset was used. \textit{Pan} stands for panoptic and \textit{Sem} for semantic segmentation.}
\resizebox{\linewidth}{!}{
    \setlength{\tabcolsep}{2pt}
    \centering
    \begin{tabular}{lccccccc}
    \toprule
    	~& SYSU& LEVIR& EGYBCD& GVLM& CLCD& OSCD& \textit{Avg}\\ \midrule
		 None & $77.0$ & $88.2$ & $62.7$ & $76.7$ & $37.0$ & $87.5$ & $71.5$\\
		 ADE20k & $81.5$ & $\mathbf{91.0}$ & $\mathbf{83.5}$ & $\mathbf{88.9}$ & $74.5$ & $50.9$ & $78.4$\\
		 City-Sem & $\mathbf{82.1}$ & $\mathbf{91.0}$ & $83.1$ & $\mathbf{88.9}$ & $\mathbf{75.3}$ & $\mathbf{52.4}$ & $\mathbf{78.8}$\\
		 COCO-Pan & $81.7$ & $90.8$ & $83.3$ & $88.5$ & $74.6$ & $49.8$ & $78.1$\\
		 IN1k & $82.0$ & $\mathbf{91.0}$ & $\mathbf{83.5}$ & $88.8$ & $74.6$ & $47.6$ & $77.9$\\
		 EuroSat & $81.5$ & $\mathbf{91.0}$ & $83.4$ & $87.8$ & $74.3$ & $49.4$ & $77.9$\\
    \bottomrule
 \end{tabular}
    }
    \label{tab:dataset}
\end{table}

\subsection{Backbone architecture}

\noindent ResNet remains a widely used backbone~\cite{chen2021bit,mall2023caco,manas2021seco}, but newer architectures offer potential improvements. We compare other common options: Swin~\cite{liu2021swin}, SwinV2~\cite{liu2022swinv2}, ViT~\cite{dosovitskiy2021vit}, ResNet18~\cite{he2016deep}, ResNet50~\cite{he2016deep}, and ConvNext~\cite{liu2022convnet}. All architectures were pre-trained on ImageNet1k for the classification task (more details in Supplementary S3-C). The results in \Cref{tab:arch} reveal a significant influence of the choice of backbone architecture on overall performance.

\begin{table}[!h]
\caption{Comparison of different backbone architectures.}

    \resizebox{\linewidth}{!}{
        \setlength{\tabcolsep}{2pt}
        \centering
        \begin{tabular}{lccccccc}
        \toprule
            ~& SYSU& LEVIR& EGYBCD& GVLM& CLCD& OSCD& \textit{Avg}\\ \midrule
             
             Swin T & $82.0$ & $\mathbf{91.0}$ & $83.5$ & $88.8$ & $74.6$ & $\mathbf{47.6}$ & $\mathbf{77.9}$\\
             SwinV2 T & $81.4$ & $\mathbf{91.0}$ & $\mathbf{84.1}$ & $89.2$ & $\mathbf{75.0}$ & $44.5$ & $77.5$\\
             ViT T & $\mathbf{82.5}$ & $84.5$ & $81.3$ & $88.0$ & $73.5$ & $40.3$ & $75.0$\\
             ViT B & $81.9$ & $85.2$ & $82.1$ & $87.7$ & $72.8$ & $42.5$ & $75.4$\\
             \midrule
             ResNet18 & $80.1$ & $89.9$ & $82.3$ & $88.9$ & $69.9$ & $40.9$ & $75.3$\\
             ResNet50 & $80.8$ & $90.3$ & $82.7$ & $\mathbf{89.5}$ & $72.5$ & $37.8$ & $75.6$\\
             ConvNext B & $81.5$ & $90.7$ & $83.7$ & $88.8$ & $73.0$ & $40.2$ & $76.3$\\
        \bottomrule
        \end{tabular}
        }
        \label{tab:arch}
    \end{table}
We hypothesise that Swin outperforms others due to its small patch size and hierarchical design. This preserves high-resolution features but still enables efficient processing of global context~\cite{liu2021swin}.

\subsection{Model size}

\noindent Larger backbones generally improve results in vision tasks given sufficient data and training~\cite{liu2021swin, dosovitskiy2021vit}. We validate this claim for the case of change detection by utilising different Swin backbone sizes (specifics in Supplementary S2-C). Results in \Cref{tab:size} show that the larger Swin-B achieves the highest performance, surpassing Swin-T by 0.8 p.\ p.\ on average. This improvement comes with almost double computational and memory costs, which could be acceptable in some critical applications.

\begin{table}[!h]
    \caption{Comparison of different Swin backbone sizes: Tiny, Small and Base.}

\resizebox{\linewidth}{!}{
    \setlength{\tabcolsep}{2pt}
    \centering
    \begin{tabular}{lccccccc}
    \toprule
    	~& SYSU& LEVIR& EGYBCD& GVLM& CLCD& OSCD& \textit{Avg}\\ \midrule
		 Swin T & $\mathbf{82.0}$ & $91.0$ & $83.5$ & $88.8$ & $74.6$ & $\mathbf{47.6}$ & $77.9$\\
		 Swin S & $81.8$ & $91.1$ & $84.1$ & $89.2$ & $75.8$ & $45.0$ & $77.8$\\
		 Swin B & $\mathbf{82.0}$ & $\mathbf{91.2}$ & $\mathbf{84.4}$ & $\mathbf{89.4}$ & $\mathbf{76.3}$ & $47.3$ & $\mathbf{78.4}$\\

    \bottomrule
    \end{tabular}
    }
    \label{tab:size}
\end{table}

\section{Training Design Choice Analysis}
\label{sec:train_opt}

\noindent After investigating the backbone design choices, we now analyse common training techniques, including augmentations, learning rate schedulers, and different loss functions. 

\subsection{Augmentations}

\begin{figure}[!h]
    \centering
    \includegraphics[width=0.9\linewidth]{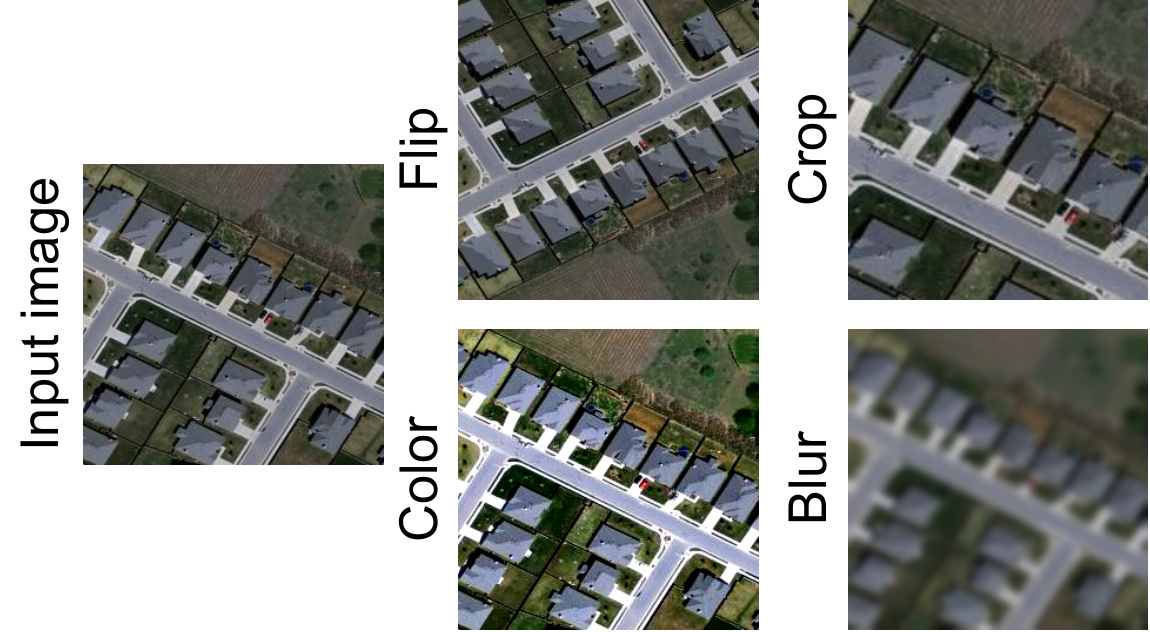}
    \caption{Illustration of different augmentations used during the analysis. The leftmost image is the unaltered input image, while all others represent augmentations of the input image.}
    \label{fig:aug}
\end{figure}

\noindent Augmentations are widely used in computer vision and, in cases such as ImageNet classification, almost standardized~\cite{liu2021swin}. However, change detection lacks such consistency, so augmentation choices vary significantly.
To assess their impact, we evaluate four common types: \textit{Flip}, \textit{Crop}, \textit{Color jitter} (named Color in \Cref{tab:augs}), and \textit{Blur}, along with some promising combinations. \Cref{fig:aug} illustrates these augmentations. All augmentations have a $30~\%$ chance of being applied and are consistent for both images in a pair.
Detailed augmentation parameters are provided in Supplementary S3-D. 

The results in \Cref{tab:augs} show that \textit{Flip} and \textit{Crop} improve performance (1.7 p.\ p.\ and 1.4 p.\ p.\ respectively). Their combination (\textit{Flip \& Crop}) also improves the baseline performance, but is less effective than only the \textit{Flip} augmentation, likely due to the excessive distortion. An exception when using \textit{Flip} occurs with the SYSU dataset, where dataset patches are already rotated beforehand~\cite{shi2022sysuDSAMnet}, reducing the benefits of this augmentation.

Although \textit{Color jitter} and \textit{Blur} augmentations are sometimes used in related methods~\cite{wang2024mtp, chen2021bit, zheng2023changen}, our findings suggest that they do not widely help with performance.

\begin{table}[!h]
        \caption{Comparison of different augmentation techniques used during training. \textit{Flip} and \textit{Crop} serve as data expanders, while \textit{Color} and \textit{Blur} can be used as countermeasures for photographic variations in data.}
\resizebox{\linewidth}{!}{
    \setlength{\tabcolsep}{2pt}
    \centering
    \begin{tabular}{lccccccc}
    \toprule
    	~& SYSU& LEVIR& EGYBCD& GVLM& CLCD& OSCD& \textit{Avg}\\ \midrule
		 None & $\mathbf{82.0}$ & $91.0$ & $83.5$ & $88.8$ & $74.6$ & $47.6$ & $77.9$\\
		 Flip & $81.3$ & $\mathbf{91.4}$ & $\mathbf{85.4}$ & $\mathbf{90.0}$ & $78.7$ & $\mathbf{51.0}$ & $\mathbf{79.6}$\\
		 Crop & $81.8$ & $91.1$ & $85.2$ & $89.3$ & $77.4$ & $50.9$ & $79.3$\\
		 Flip \& Crop & $80.3$ & $\mathbf{91.4}$ & $84.8$ & $\mathbf{90.0}$ & $\mathbf{79.3}$ & $49.1$ & $79.2$\\
		 Color & $81.7$ & $91.0$ & $83.7$ & $89.0$ & $74.4$ & $47.9$ & $77.9$\\
		 Blur & $81.4$ & $91.1$ & $83.6$ & $88.9$ & $74.8$ & $46.5$ & $77.7$\\
    \bottomrule
    \end{tabular}
    }

    \label{tab:augs}
\end{table}

We hypothesise that \textit{Flip} and \textit{Crop} play a similar role during training -- extending the dataset. Since remote sensing data lacks a fixed orientation~\cite{rolf2024position}, \textit{flipping} is a natural choice. The \textit{Crop} augmentation also plays an important role, forcing the model to adapt to variations in ground sample distance~\cite{tang2024xscalemae}. In contrast, the \textit{Color jitter} and \textit{Blur} augmentations appear to be more dataset-dependent, potentially serving as a countermeasure for photographic variations in images (e.g. brightness, contrast, blur, etc.).

\subsection{Learning rate scheduler}

\noindent Learning rate schedulers are widely used to stabilise model training and improve performance, but no single choice dominates in change detection. We thus evaluate five common types: \textit{Multistep}, \textit{Cosine}, \textit{Exponential}, \textit{Linear} and \textit{Polynomial}. Warm-up variants were also tested but did not yield improvements, so they are omitted. Implementation details are in Supplementary S3-E.

As shown in \Cref{tab:sched}, training without a scheduler achieves the best result. However, as we will later show (\Cref{tab:combined}), the \textit{Cosine}~\cite{loshchilov2017sgdr} scheduler proves beneficial when combined with augmentations. Our intuition is that it serves as a countermeasure towards the end of training for increased variability introduced by augmentations. 

\begin{table}[!ht]
    \caption{Comparison of different learning rate schedulers. The best-performing \textit{Cosine} scheduler performs worse than the baseline, but as we show in \Cref{tab:combined}, it helps when combined with random augmentations.}
\resizebox{\linewidth}{!}{
    \setlength{\tabcolsep}{2pt}
    \centering
    \begin{tabular}{lccccccc}
    \toprule
    	~& SYSU& LEVIR& EGYBCD& GVLM& CLCD& OSCD& \textit{Avg}\\ \midrule
		 None & $82.0$ & $\mathbf{91.0}$ & $\mathbf{83.5}$ & $88.8$ & $\mathbf{74.6}$ & $\mathbf{47.6}$ & $\mathbf{77.9}$\\
		 Multistep & $81.9$ & $90.8$ & $82.8$ & $\mathbf{88.9}$ & $73.5$ & $43.6$ & $76.9$\\
		 Cosine & $81.9$ & $90.4$ & $82.9$ & $88.5$ & $73.5$ & $45.3$ & $77.1$\\
		 Exp. & $\mathbf{82.2}$ & $89.4$ & $81.9$ & $88.2$ & $72.7$ & $46.0$ & $76.7$\\
		 Linear & $81.9$ & $90.3$ & $82.8$ & $88.6$ & $73.2$ & $44.6$ & $76.9$\\
		 Poly. & $81.8$ & $90.4$ & $82.8$ & $88.7$ & $73.3$ & $44.3$ & $76.9$\\

    \bottomrule
    
    \end{tabular}
    }

    \label{tab:sched}
\end{table}

\subsection{Loss}

\noindent Cross-entropy (CE) loss is the most widely used loss function in related works, but alternative choices, such as Dice loss~\cite{Milletari2016diceVnet} and Focal loss~\cite{lin2017focal}, have been introduced in recent methods. These are often combined with CE, so we evaluate both individual and combined variants and present the results in \Cref{tab:loss}.

\begin{table}[!ht]
    \caption{Comparison of different loss functions on change detection performance.}

\resizebox{\linewidth}{!}{
    \setlength{\tabcolsep}{2pt}
    \centering
    \begin{tabular}{lccccccc}
    \toprule
    	~& SYSU& LEVIR& EGYBCD& GVLM& CLCD& OSCD& \textit{Avg}\\ \midrule
		 CE & $\mathbf{82.0}$ & $91.0$ & $83.5$ & $88.8$ & $74.6$ & $47.6$ & $77.9$\\
		 Focal & $81.8$ & $90.9$ & $83.1$ & $88.9$ & $\mathbf{75.2}$ & $44.1$ & $77.3$\\
		 Dice & $81.7$ & $91.0$ & $83.3$ & $\mathbf{89.0}$ & $74.7$ & $\mathbf{53.0}$ & $\mathbf{78.8}$\\
		 Focal+Dice & $81.5$ & $90.9$ & $\mathbf{83.8}$ & $\mathbf{89.0}$ & $74.4$ & $51.0$ & $78.4$\\
		 CE+Dice & $81.8$ & $\mathbf{91.1}$ & $83.7$ & $\mathbf{89.0}$ & $74.1$ & $48.0$ & $77.9$\\

    \bottomrule
    \end{tabular}
    
    }
    \label{tab:loss}
\end{table}

While performance varies slightly across datasets, \textit{Dice} loss achieves the best overall result, particularly on the low-resolution OSCD dataset. Dice loss is well-suited for unbalanced data~\cite{sudre2017genDice} and is usually used as an auxiliary function in related work. However, our results suggest that it is very effective as a standalone primary loss function.

\section{Putting It All Together}
\label{sec:sota}
\label{sub:together}

\noindent This section combines all the findings from our analysis to create a single optimised method called \textit{BTC} (\textbf{B}e \textbf{T}he \textbf{C}hange). We present the results when gradually adding improvements to the method, and then show that our insights also apply to related methods. Lastly, we compare \textbf{BTC} to state-of-the-art approaches.

To ensure practical comparisons while avoiding infeasible amounts of experiments, we incrementally refine the model by applying some of the strongest design choices from each component and report the impact of the best option. 

\begin{table}[!h]
     \caption{Results of combining our insights to reach the final optimised change detection architecture. Each row presents the F1 score and the difference compared to the previous row. Standard deviation is presented in Supplementary S4. "+" indicates introduction of new design choices, while \switch{} represents the exchange of one design choice with another.}
\resizebox{\linewidth}{!}{
    \setlength{\tabcolsep}{2pt}
    \centering
    \begin{tabular}{lccccccc}
    \toprule
    	~& SYSU& LEVIR& EGYBCD& GVLM& CLCD& OSCD& \textit{Avg}\\ \midrule
		 \rowcolor{gray!20} Swin-T & $77.0 $ & $88.2 $ & $76.7 $ & $87.5 $ & $62.7 $ & $37.0 $ & $71.5 $\\
          \midrule
		 + \makecell[l]{IN1k \\ pre-train} &  \resInc{$82.0$} {5.0} &  \resInc{$91.0$} {2.8} &  \resInc{$83.5$} {6.8} &  \resInc{$88.8$} {1.3} &  \resInc{$74.6$} {11.9} &  \resInc{$47.6$} {10.6} &  \resInc{$77.9$} {6.4}\\
		 \midrule 
		 + \makecell[l]{Flip \\ augment} &  \resDec{$81.3$} {0.7} &  \resInc{$91.4$} {0.4} &  \resInc{$85.4$} {1.9} &  \resInc{$90.0$} {1.2} &  \resInc{$78.7$} {4.1} &  \resInc{$51.0$} {3.4} &  \resInc{$79.6$} {1.7}\\
		 \midrule 
		 \switch{} \makecell[l]{CityS. \\ pre-train} &  \resDec{$81.0$} {0.3} &  \resInc{$91.5$} {0.1} &  \resInc{$85.9$} {0.5} &  \resInc{$90.1$} {0.1} &  \resInc{$79.1$} {0.4} &  \resInc{$54.8$} {3.8} &  \resInc{$80.4$} {0.8}\\
		 \midrule 
		 + \makecell[l]{Cosine \\ scheduler} & \resSame{$81.0$} &  \resInc{$91.6$} {0.1} &  \resInc{$\mathbf{86.2}$} {0.3} &  \resInc{$90.5$} {0.4} &  \resInc{$79.4$} {0.3} &  \resInc{$\mathbf{54.9}$} {0.1} &  \resInc{$80.6$} {0.2}\\
		 \midrule 
		 \switch{} Swin-B &  \resInc{$81.8$} {0.8} &  \resInc{$\mathbf{91.7}$} {0.1} &  \resDec{$86.0$} {0.2} &  \resInc{$\mathbf{90.7}$} {0.2} &  \resInc{$\mathbf{81.6}$} {2.2} &  \resDec{$52.4$} {2.5} &  \resInc{$80.7$} {0.1}\\
		 \midrule 
		 \switch{} Dice loss &  \resInc{$\mathbf{82.4}$} {0.6} &  \resDec{$91.5$} {0.2} &  \resDec{$85.6$} {0.4} & \resSame{$90.7$} &  \resDec{$80.9$} {0.7} &  \resInc{$54.3$} {1.9} &  \resInc{$\mathbf{80.9}$} {0.2}\\

    \bottomrule
    \end{tabular}
    }

    \label{tab:combined}
\end{table}
We begin with a randomly initialised Swin-T trained with cross-entropy loss, no augmentations, and no schedulers. The most significant improvement comes from pre-training: introducing \textit{ImageNet1k} pre-trained weights boosts performance by $6.4$ p.p. Adding \textit{flip} augmentations provides an additional $1.7$ p.p. gain. Switching from an ImageNet1k to a \textit{Cityscapes} semantic segmentation pre-trained backbone contributes another $0.8$ p.p. We also evaluate the \textit{cosine} scheduler since it is the best non-default option from the category. Although it shows no benefits in isolation, it contributes $0.2$ p.p when combined with augmentations and different pre-training. Lastly, increasing the backbone size to \textit{Swin-B} and replacing CE loss with \textit{Dice} loss yields gains of $0.1$ p.p. and $0.2$ p.p., respectively. 

Together, these optimisations raise the average performance by $9.4$ p.p., from $71.5~\%$ to $80.9~\%$, with total gains up to $18.2$ and $17.3$ p.p. on CLCD and OSCD, respectively. The results highlight the impact of basic design choices that can be easily overlooked or inconsistently applied. The substantial improvements from pre-training type selection and simple augmentation show that models can underperform due to suboptimal defaults rather than inherent architectural limitations. This underscores the need for rigorous baseline evaluations to ensure future change detection research builds on meaningful advancements.

\begin{figure}[!h]
    \centering
    \includegraphics[width=0.6\linewidth]{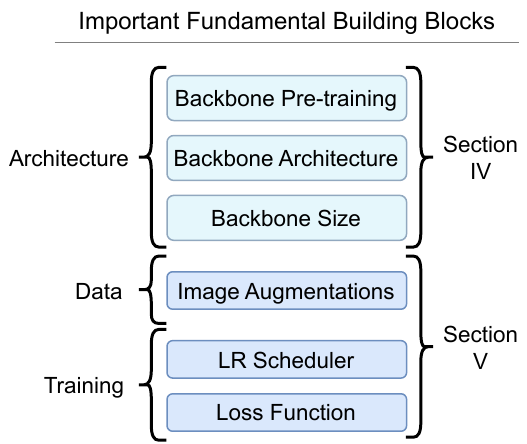}
    \caption{\change{Summary of important fundamental building blocks analysed in our study. We consider various architectural, data, and training aspects. The right side of the figure indicates in which section we analyse a specific element: architectural aspects in \Cref{sec:bbone_opt}, and data with training aspects in \Cref{sec:train_opt}.}}
    \label{fig:summary}
\end{figure}

\begin{figure*}[!t]
    \centering
    \includegraphics[width=1\linewidth]{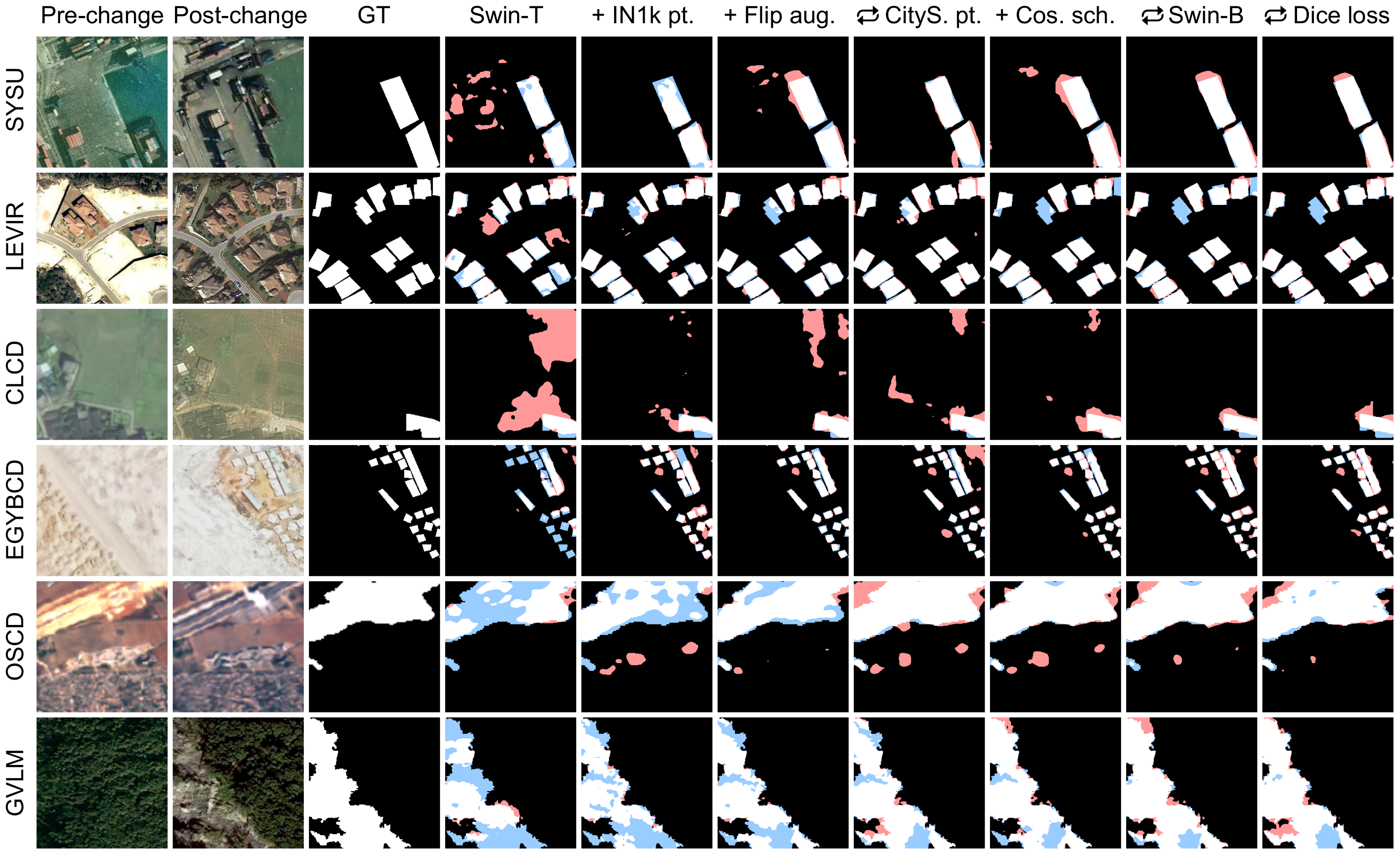}
    \caption{Qualitative comparison of the predictions made by models with different combined configurations. The pair of considered images is shown in the first and second columns, followed by the ground truth mask in the third column. The next seven columns depict predictions. \textcolor{falsePos}{False positives are marked in red} and \textcolor{falseNeg}{false negatives in blue}. "+" indicates the introduction of a new design choice, while \switch{} represents the exchange of an existing design choice. Additional qualitative results are in Supplementary S5}
    \label{fig:qual}
    
\end{figure*}

We further analyse our approach through qualitative results in \Cref{fig:qual}.
While performance usually improves with additions, some intermediate steps cause temporary deterioration. Compared to the initial model, the final version better discriminates between irrelevant changes and changes of interest, reducing false positives (\textcolor{falsePos}{red} regions). It also improves differentiation of visually or semantically similar changes, leading to fewer false negatives (\textcolor{falseNeg}{blue} regions).

\change{A visual summary of the considered elements is presented in \Cref{fig:summary}. Based on the performed analysis, the quantitative results of combined insights from \Cref{tab:combined}, and qualitative results from \Cref{fig:qual}, it can be concluded that each fundamental element plays an important role in achieving the final performance. Among the most influential are the selection of \textbf{backbone architecture and pre-training dataset}, and the \textbf{data augmentations} used. When designing change detection models, selecting a strong backbone like Swin, with high spatial resolution and simultaneous global perception, is important. This backbone also greatly benefits from suitable pretraining similar to the change detection task. Due to the imbalance problem of change detection, augmentations that expand the dataset, like flipping and rotating, also creatively contribute to the final result. Finally, it is also important to consider other training elements, among which the loss function and learning rate scheduler play an important role.}

\subsection{Transfer to related methods}
\label{sec:transfer}

\noindent We applied the design choices to related methods to showcase their broader applicability. The results are presented in \Cref{tab:sota_pt}.
\begin{table}[!h]
    \caption{Average F1 across 6 datasets for different SOTA methods with and without using optimisations from our insights (full results in Supplementary S3-G and standard deviation in Supplementary S4-A).}
\resizebox{\linewidth}{!}{
    \setlength{\tabcolsep}{2pt}    
    \centering
    \begin{tabular}{lcccccccc}
    \toprule
    	~& \makecell{SeCo \\ ~\cite{manas2021seco}}& \makecell{CaCo \\ ~\cite{mall2023caco}}& \makecell{SatMAE \\ ~\cite{cong2022satmae}}& \makecell{GFM \\ \cite{mendieta2023gfm}}& \makecell{GeSSL \\ ~\cite{ayush2021gessl}}& \makecell{MTP \\ \cite{wang2024mtp}}& \makecell{FCSiamD\\\cite{daudt2018fcn}}& \makecell{SwinSUN\\\cite{zhang2022swinsunet}}\\ \midrule
		 \rowcolor{gray!20} Base & $74.0$ & $75.7$ & $77.1$ & $77.7$ & $75.9$ & $77.2$ & $60.5$ & $78.0$\\
		 + Opt. &  \resInc{$77.0$} {\footnotesize3.0} &  \resInc{$79.3$} {\footnotesize3.6} &  \resInc{$79.5$} {\footnotesize2.4} &  \resInc{$79.6$} {\footnotesize1.9} &  \resInc{$79.7$} {\footnotesize3.8} &  \resInc{$80.3$} {\footnotesize3.1} &  \resInc{$77.4$} {\footnotesize16.9} &  \resInc{$79.6$} {\footnotesize1.6}\\

    \bottomrule
    \end{tabular}
    }
    \label{tab:sota_pt}
\end{table}
Each remote sensing foundation model (SeCo, CaCo, SatMAE, GFM, GaSSL, and MTP) is integrated as the backbone within our framework (\Cref{sec:method}). The performance is first assessed under the base configuration: no augmentations, no scheduler, CE loss for foundation models, and the author's default configuration for change-detection-specific models. Additional architecture and hyperparameter details are in Supplementary S3-F. We then use flip augmentations, Dice loss, and a cosine scheduler. For change-detection-specific models (FC-Siam-Diff~\cite{daudt2018fcn} and SwinSUNet~\cite{zhang2022swinsunet}) we also exchange their default encoder with Cityscapes pre-trained Swin-B used by our approach. This leads to consistent performance gains of several percentage points (full results are in Supplementary S3-G).

The large improvement of $16.9$ p.p for FC-Siam-Diff and noticeable increase of $1.6$ p.p. for SwinSUNet demonstrate that fundamental design choices -- such as proper pre-training and training strategies -- were previously underutilised in change detection architectures. Despite their architectural innovations, these methods did not fully optimise base components, likely due to the absence of systematic analyses. This highlights the value of our findings in guiding future research toward more effective and well-founded model design.

\begin{table*}[!h]
    \caption{Comparison of the proposed method to the state-of-the-art on six different datasets. We report the mean F1 of three runs with different random seeds for each dataset and an average with standard deviation across all datasets (standard deviation for individual datasets is in Supplementary S4-B). \textcolor{goldD}{First}, \textcolor{silverD}{second} and \textcolor{bronzeD}{third} place results are marked. \colorbox{TealBlue!15}{Remote sensing pre-training based} methods are highlighted in \colorbox{TealBlue!15}{blue}, while \colorbox{orange!15}{change detection specific} architectures are marked with \colorbox{orange!15}{orange}. FPS was benchmarked on an Nvidia A100-SXM4 40GB GPU (Details in Supplementary S3-H).}
 \resizebox{\linewidth}{!}{
    \setlength{\tabcolsep}{2pt}
    \centering
    \begin{tabular}{lcccc|ccccccc}
    \toprule
    	~ & Backbone & Pre-train & FPS [img/s] & Param. [M] & SYSU& LEVIR& EGYBCD& GVLM& CLCD& OSCD& \textit{Avg \& std}\\ \hline
		 \cellcolor{orange!15}FCS-Diff~\cite{daudt2018fcn} & UNet & - & $170.1$ & $1.4$ & $70.8$ & $81.8$ & $42.3$ & $74.3$ & $54.1$ & $39.4$ & $60.5${ \scriptsize $\pm0.5$ }\\
		 \cellcolor{orange!15}BIT~\cite{chen2021bit} & RN-18 & IN1k & $57.7$ & $12.4$ & $77.0$ & $90.0$ & $75.9$ & $88.8$ & $63.4$ & $42.6$ & $73.0${ \scriptsize $\pm0.4$ }\\
		 \cellcolor{orange!15}ChangeFormer~\cite{bandara2022changeFormer} & MiT-B0 & - & $36.2$ & $41.0$ & $77.9$ & $89.5$ & $77.9$ & $88.5$ & $60.8$ & $48.1$ & $73.8${ \scriptsize $\pm0.3$ }\\
		 \cellcolor{orange!15}BiFA~\cite{zhang2024bifa} & MiT-B0 & ADE20k & $32.2$ & $9.9$ & \bm1{83.8} & $89.5$ & $83.5$ & $89.0$ & $74.5$ & $37.4$ & $76.3${ \scriptsize $\pm0.8$ }\\
		 \cellcolor{TealBlue!15}SeCo~\cite{manas2021seco} & RN-50 & RS & $100.7$ & $64.0$ & $78.9$ & $90.6$ & $84.1$ & \bm3{90.4} & $68.7$ & $49.1$ & $77.0${ \scriptsize $\pm0.2$ }\\
		 \cellcolor{orange!15}SwinSUNet~\cite{zhang2022swinsunet} & Swin-T & IN1k & $33.1$ & $43.6$ & $76.6$ & $89.3$ & $83.7$ & $90.0$ & $75.8$ & $52.8$ & $78.0${ \scriptsize $\pm0.4$ }\\
		 \cellcolor{TealBlue!15}CaCo~\cite{mall2023caco} & RN-50 & RS & $100.9$ & $64.0$ & $80.2$ & $90.9$ & $84.8$ & $90.3$ & $77.9$ & $51.9$ & $79.3${ \scriptsize $\pm0.0$ }\\
		 \cellcolor{TealBlue!15}SatMAE~\cite{cong2022satmae} & ViT-L & RS & $71.2$ & $322.6$ & $81.4$ & \bm3{91.4} & \bm1{85.9} & $89.8$ & $79.2$ & $49.5$ & $79.5${ \scriptsize $\pm0.2$ }\\
		 \cellcolor{TealBlue!15}GFM~\cite{mendieta2023gfm} & Swin-B & RS & $44.9$ & $120.5$ & $81.2$ & $89.8$ & $85.0$ & $89.8$ & $77.5$ & \bm2{54.1} & $79.6${ \scriptsize $\pm0.3$ }\\
		 \cellcolor{TealBlue!15}GeSSL~\cite{ayush2021gessl} & RN-50 & RS & $100.4$ & $64.0$ & $80.2$ & $91.2$ & \bm3{85.4} & $90.2$ & $78.1$ & \bm3{53.1} & $79.7${ \scriptsize $\pm0.2$ }\\
		 \cellcolor{TealBlue!15}MTP~\cite{wang2024mtp} & ViT-B + RVSA & RS & $31.2$ & $107.8$ & $81.3$ & \bm1{91.7} & \bm1{85.9} & $89.9$ & \bm2{80.3} & $52.8$ & \bm3{80.3}{ \scriptsize $\pm0.1$ }\\
		 BTC {\footnotesize T} (Ours) & Swin-T & CityS. & $57.8$ & $58.9$ & \bm3{81.6} & \bm3{91.4} & \bm2{85.6} & \bm2{90.6} & \bm3{79.5} & \bm1{54.3} & \bm2{80.5}{ \scriptsize $\pm0.2$ }\\
		 \textbf{BTC {\footnotesize B}} (Ours) & Swin-B & CityS. & $32.4$ & $120.1$ & \bm2{82.4} & \bm2{91.5} & \bm2{85.6} & \bm1{90.7} & \bm1{80.9} & \bm1{54.3} & \bm1{80.9}{ \scriptsize $\pm0.0$ }\\
         
         \bottomrule
    \end{tabular}
    }
    \label{tab:sota}
\end{table*}

\subsection{Comparison with state-of-the-art}

\noindent To showcase the potency of our findings, we compared BTC to the current state-of-the-art. We include the following remote sensing foundation models: SatMAE~\cite{cong2022satmae}, GeSSL~\cite{ayush2021gessl}, CaCo~\cite{mall2023caco}, SeCo~\cite{manas2021seco}, MTP~\cite{wang2024mtp}, and GFM~\cite{mendieta2023gfm}.
We also include methods \textit{specifically designed for change detection}: 
BIT~\cite{chen2021bit}, FC-Siam-Diff~\cite{daudt2018fcn}, ChangeFormer~\cite{bandara2022changeFormer}, SwinSUNet~\cite{zhang2022swinsunet}, and BiFA~\cite{zhang2024bifa}. Since remote sensing foundation models don't have a specific training protocol, we utilise our optimised framework. Change detection-specific architectures already define the training protocol, so we use their original setup.

In accordance with our protocol, we rerun all experiments with three different random seeds and report the average result. We also report standard deviations for dataset averages here, while complete deviation results are in Supplementary S4-B and implementation details in Supplementary S3-F.

\subsubsection{Change detection performance}

\Cref{tab:sota} presents a comparison of our approach to state-of-the-art methods. By leveraging all insights from our analysis, BTC surpasses previously proposed models on the CLCD, GVLM, and OSCD datasets. It also achieves competitive results on SYSU, LEVIR, and EGYBCD and achieves the best average performance across datasets. 

\textit{BTC T} outperforms SwinSUNet~\cite{zhang2022swinsunet} by $2.5$ p.p. using the same Swin-T backbone, a simpler fusion module and decoder, but a more optimal training setup. This reiterates our claim that previous methods have not sufficiently explored the base design choices. The gap is reduced to $1.3$ p.p. if we apply our optimisations to SwinSUNet (presented in \Cref{sec:transfer}). This means BTC still performs better, suggesting that some complex fusion and decoder blocks might even harm the performance when paired with optimal training setups.

Remote sensing pre-trained models generally outperform change detection methods on smaller datasets like CLCD and OSCD, where limited data hinders the training of architecturally advanced fusion and decoder modules. Yet, our approach outperforms even foundation models despite using weights pre-trained on natural images. As discussed in \Cref{subsec:pt-data} and in~\cite{sosa2024effective}, this highlights the importance of task selection in pre-training, not just representation quality. Segmentation pre-trained weights benefit change detection, which may also explain the strong performance of MTP~\cite{wang2024mtp}, where segmentation is one of the pre-training tasks.

\subsubsection{Computational efficiency}

While our approach and foundation models have larger parameter sizes, their inference speed (measured in frames per second -- FPS -- 4th column in \Cref{tab:sota}) can still match smaller architectures on efficient modern hardware. Details on the computational efficiency evaluation protocol with additional metrics are in Supplementary S3-H.

\subsubsection{Qualitative results}

\begin{figure}[!h]
    \centering
    \includegraphics[width=1\linewidth]{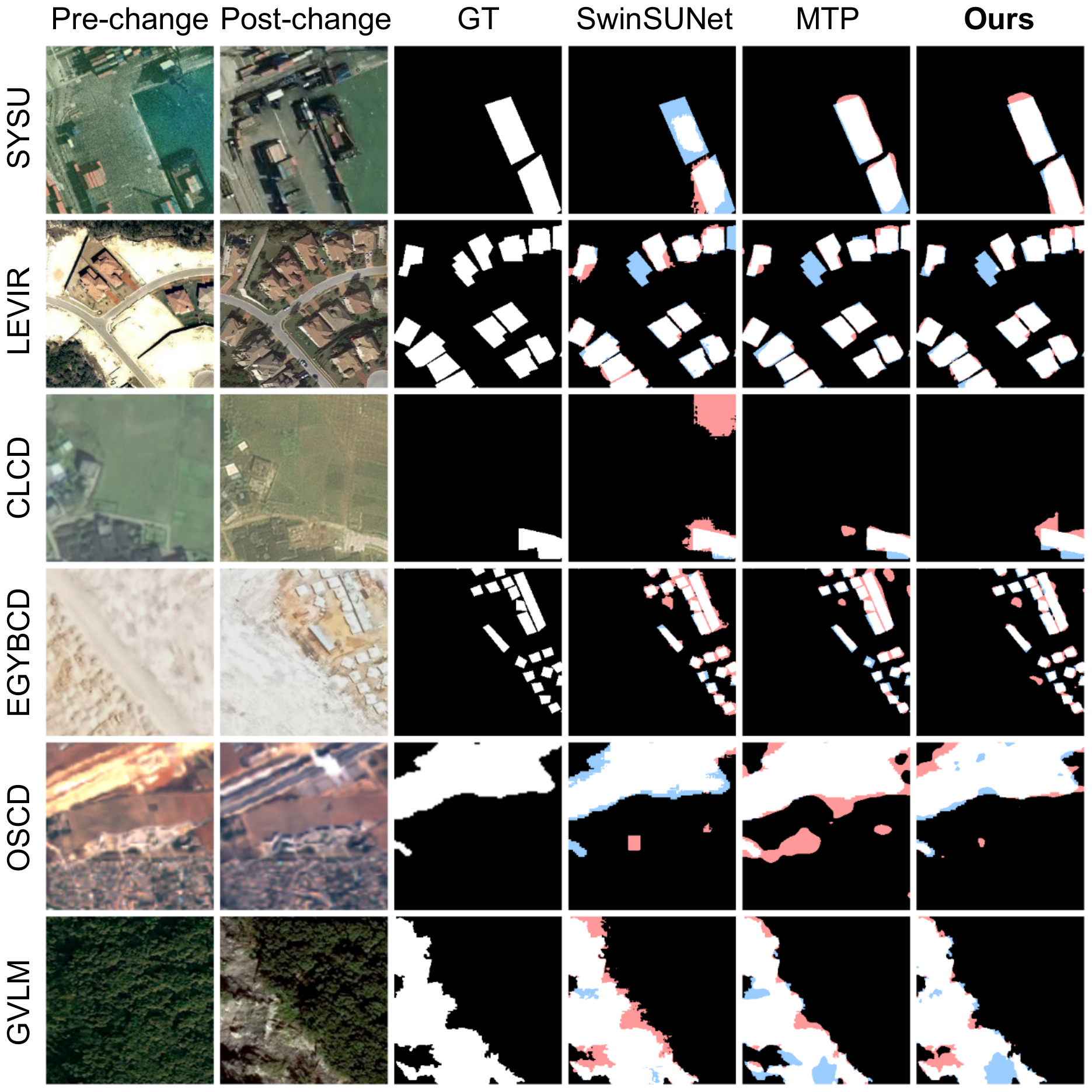}
    \caption{Qualitative comparison of the predictions made by our method and best-performing related methods: SwinSUNet (change-detection-specific) and MTP (foundation model). The pair of considered images is shown in the first and second columns, followed by the ground truth mask and predictions for each method. \textcolor{falsePos}{False positives are marked in red} and \textcolor{falseNeg}{false negatives in blue}. Qualitative results for all other models are in Supplementary S5-B.}
    \label{fig:sota}
\end{figure}

Qualitative results in \Cref{fig:sota} show predictions for our method compared to the best-performing remote sensing foundation model MTP~\cite{wang2024mtp} and the best-performing change detection architecture SwinSUNet~\cite{zhang2022swinsunet}. While all methods predict visually similar results, our method, in some cases, produces fewer \textcolor{falsePos}{false positive} regions.

\subsection{Discussion}

The results emphasise the critical role of carefully selecting fundamental elements of change detection methods. While the optimisations we highlight may appear standard, their application remains inconsistent, likely due to their perceived simplicity. Improved basic and standard elements potentially overshadow other architectural additions in new methods because their impact was not well established. This also raises the question of how much further the performance could be improved with advancements built with \change{better} fundamental components. 

BTC and the accompanying analysis provide a strong and transparent baseline for both change detection architectures and remote sensing pre-training. Evaluation of future methods could benefit from clarity, preventing situations where novel architectural additions potentially reduce the performance compared with the baseline, utilising the same backbone but simpler fusion and decoder architecture. The results also strengthen the need for pre-training strategies tailored to the task --potentially leveraging segmentation-inspired techniques-- while integrating remote-sensing-specific properties to outperform natural-image segmentation pre-training. 

\change{While our final optimised model employs a Swin architecture, we recognise that alternative architectures might exhibit different optimal configurations concerning various design options. Nevertheless, our study emphasises the substantial influence of configurations such as pretraining and augmentations (\Cref{tab:combined}). We anticipate that future research will take these aspects into account more diligently, as they were not sufficiently explored before.}

\section{Conclusion}
\label{sec:conclusion}

In this work, we present BTC, a simple yet strong model for change detection, which achieves state-of-the-art performance across six datasets. Our approach highlights the importance of fundamental design choices that drive performance, supported by an extensive empirical analysis. By analysing backbone configurations and training strategies, we offer valuable insights and guidelines for future methods. The impact of optimisations resulting from our analysis is evident in the base model’s performance improvement of F1 from $71.5~\%$ to $80.9~\%$ across 6 diverse change detection datasets. 

We further demonstrate the impact of our observations by applying them to related models, significantly improving their performance. The results suggest that previous approaches have often overlooked the contribution of these fundamental choices, potentially due to a lack of a detailed analysis. We thus believe that our findings and the BTC model lay a strong foundation for future advancements in the field.

A key direction for future work is extending these insights to multi-modal data and other remote sensing tasks. This includes adapting pre-trained weights for SAR, multispectral, and hyperspectral applications and refining pre-training strategies to better align with change detection and other dense prediction tasks, such as segmentation.

\section*{Acknowledgments}
This work was in part supported by the ARIS research project J2-60045, research programme P2-0214, and the supercomputing network SLING (ARNES, EuroHPC Vega). 

\bibliography{main}
\bibliographystyle{IEEEtran}

\clearpage
\makeatletter
\renewcommand \thesection{S\@arabic\c@section}
\renewcommand\thetable{S\@arabic\c@table}
\renewcommand \thefigure{S\@arabic\c@figure}
\makeatother

\setcounter{page}{1}
\setcounter{footnote}{0}
\setcounter{table}{0}
\setcounter{figure}{0}
\setcounter{section}{0}
\maketitlesupplementary

\section{Additional Dataset Information}
\label{a:data}

In this section, we provide additional details about the datasets used in our experiments. Technical specifications are covered in \Cref{asub:data_details}, with implementation details in \Cref{asub:data_impl}, including splits and download sources.

\subsection{Dataset Details}
\label{asub:data_details}

Extended dataset specifications are listed in \Cref{atab:data_extend}. The datasets cover a wide range of change types, resolutions (GSDs), and acquisition sensors. They also vary in size, ranging from a few hundred to several thousand image pairs. This diversity enhances the robustness and general applicability of our findings. 

Due to the small percentage of changed pixels (typically below $10~\%$), remote sensing change detection remains highly imbalanced and challenging. The higher change ratio in SYSU is primarily due to coarse annotations for larger change regions.

\subsection{Implementation Details}
\label{asub:data_impl}

\textbf{Splits} Official splits are used for OSCD, SYSU, CLCD, and LEVIR, while EGY-BCD and GVLM rely on already available public random splits from Huggingface (courtesy of Weikang Yu) to ensure full reproducibility and fair comparison.

The links for data used are as follows:
\begin{itemize}
    \item SYSU: \url{https://huggingface.co/datasets/ericyu/SYSU_CD}
    \item LEVIR: \url{https://huggingface.co/datasets/ericyu/LEVIRCD_Cropped256}
    \item EGY-BCD: \url{https://huggingface.co/datasets/ericyu/EGY_BCD}
    \item GVLM: \url{https://huggingface.co/datasets/ericyu/GVLM_Cropped_256}
    \item CLCD: \url{https://huggingface.co/datasets/ericyu/CLCD_Cropped_256}
    \item OSCD: \url{https://rcdaudt.github.io/oscd/}. Since this dataset is not prepared, we manually crop the images from predefined train and test split into correctly shaped patches. For easier future use, we will release our hdf5 split files upon acceptance.
\end{itemize}
For details on normalisation, resizing and random augmentations, refer to \Cref{asub:aug_imp}.

\section{Evaluation Protocol and Metrics}
\label{a:eval}

We use \textbf{F1-score} in our experiments to measure the change detection performance. F1 combines precision and recall into a single metrics.
For binary case, the precision and recall are calculated as follows:
\begin{equation}
\text{precision} = \frac{TP}{TP + FP}
\end{equation}

\begin{equation}
\text{recall} = \frac{TP}{TP + FN}
\end{equation}
And F1 is then the harmonic mean of precision and recall:
\begin{equation}
F_1 = 2 \cdot \frac{\text{precision} \cdot \text{recall}}{\text{precision} + \text{recall}} 
= \frac{2TP}{2TP + FP + FN}
\end{equation}
In this case, \textit{true positives} (TP) are changed pixels correctly predicted as changed, and \textit{true negatives} are unchanged pixels correctly predicted as unchanged. \textit{False positives} (FP) are unchanged pixels misclassified as changed, and \textit{false negatives} (FN) are changed pixels misclassified as unchanged.

As already discussed in the main paper, the proper F1 calculation in change detection follows binary formulation~\cite{bandara2022changeFormer, wang2024mtp, chen2021bit, daudt2018fcn, zheng2023changen}, where the performance of unchanged pixels classification is already reflected through TNs and FNs.

Some works have deviated from this by computing F1 separately for changed and unchanged pixels, then computing the mean F1-score: $m\mathrm{F1} = (\mathrm{F1}^c + \mathrm{F1}^{uc}) / 2$. This leads to inflation of scores due to class imbalance, preventing direct comparison of methods. To illustrate this, we evaluate our best performing model with both binary F1 and $m\mathrm{F1}$ in \Cref{tab:f1_tab}. As we can see, the incorrectly computed metric skews the results. A poorly performing method using $m\mathrm{F1}$ may appear competitive with a superior method evaluated correctly.
\begin{table}[!h]
    \caption{Comparison of binary F1-score to 2 class mean F1-score. The results are skewed and prevent direct comparison between methods.}
    \resizebox{\linewidth}{!}{
    \setlength{\tabcolsep}{2pt}
    \centering
        \begin{tabular}{lccccccc}
        \toprule
    	~& SYSU& LEVIR& EGY-BCD& GVLM& CLCD& OSCD& \textit{Avg}\\ \midrule
        F1 & 82.1 & 91.5 & 85.7 & 90.7 & 81.1 & 54.5 & 80.9 \\
        mF1 & 88.6 & 95.5 & 92.4 & 95.0 & 89.8 & 76.2 & 89.6 \\
        \bottomrule
        \end{tabular}
        }

    \label{tab:f1_tab}
\end{table}

\begin{table*}
    \caption{Additional details for the datasets used in the paper.}

\resizebox{\linewidth}{!}{
\setlength{\tabcolsep}{2pt}
    \begin{tabular}{lcccccccccc}
        \toprule
 & Acquisition & Resolution & Change Type & Interval & Region & \makecell{Image count \\ train\textbackslash val\textbackslash test} & Patch & \makecell{Changed \\ Pixels} & \makecell{Unchanged \\ Pixels} & Total Pixels \\
\midrule
SYSU~\cite{shi2022sysuDSAMnet} & Aerial & 0.5m & \makecell{Buidling, urban,\\groundwork, road,\\vegetation, sea} & 2007-2014 & Hong Kong & \makecell{12000\\4000\\4000} & $256\times256$ & $21.8~\%$ & $78.2~\%$ & 1310720000 \\\midrule
LEVIR~\cite{chen2020levirStanet} & \makecell{Google Earth \\ satellite} & 0.5m & Buidling & 2002-2018 & \makecell{20 regions \\ in US} & \makecell{7120\\1024\\2048} & $256\times256$ & $4.7~\%$ & $95.3~\%$ & 667942912 \\\midrule
EGYBCD~\cite{holail2023egybcd} & \makecell{Google Earth \\ satellite} & 0.25m & Building & 2015-2022 & Egypt & \makecell{3654\\1219\\1218} & $256\times256$ & $7.0~\%$ & $93.0~\%$ & 399179776 \\\midrule
GVLM~\cite{zhang2023gvlm} & \makecell{Google Earth \\ satellite \\ (SPOT-6)} & 0.59m & Landslide & 2010-2021 & \makecell{17 regions\\ worldwide} & \makecell{4558\\1519\\1519} & $256\times256$ & $6.6~\%$ & $93.4~\%$ & 497811456 \\\midrule
CLCD~\cite{li2022clcdMSCANET} & \makecell{Satellite \\ (Gaofen-2)} & 0.5m-2m & \makecell{Multiple types\\limited to\\croplands} & 2017-2019 & \makecell{Guangdong,\\ China} & \makecell{1440\\480\\480} & $256\times256$ & $7.6~\%$ & $92.4~\%$ & 157286400 \\\midrule
OSCD~\cite{daudt2018urban} & \makecell{Satellite \\ (Sentinel-2)} & 10m & Urban & 2015-2018 & \makecell{24 regions \\ worldwide} & \makecell{827\\-\\385} & $96\times96$ & $3.2~\%$ & $96.8~\%$ & 11169792 \\
\bottomrule
    \end{tabular}
    }
    \label{atab:data_extend}
\end{table*}

To ensure fair and direct comparison, we urge researches to correctly use binary F1 metric, focusing only on the change class. We avoid errors by using \textit{BinaryF1Score} from Torchmetrics\footnote{\url{https://lightning.ai/docs/torchmetrics/stable/classification/f1_score.html##binaryf1score}}.

\section{Extended Results and Implementation Details}

This section provides additional results and implementation details for our experiments.
We first cover backbone pre-training details and additional results in \Cref{a:pt}. Next we have implementation details for backbone architectures in \Cref{asub:arch} and Swin backbone sizes in \Cref{asub:bb_size}. Implementation details for used augmentations are in \Cref{asub:aug_imp} and in \Cref{asub:sched} for schedulers.

We then provide implementation details for related methods in \Cref{asub:relat_imp}, followed by complete results on transferring our insights to other methods in \Cref{asub:trans_ext}. Finally, we provide details on computational efficiency protocol along with additional results in \Cref{sub:comp}.

\subsection{Backbone Pre-training}
\label{a:pt}

This subsection first provides implementation details for the pre-trained weights used in our experiments (\Cref{asub:pt_impl}) and then presents results for additional pre-training tasks-dataset pairs (\Cref{asub:pt_task}).

\subsubsection{Implementation Details}
\label{asub:pt_impl}

We use pre-trained weights from Huggingface\footnote{\url{https://huggingface.co/models}} in our experiments. The corresponding Swin-T weights for each dataset-task pair are as follows:
\begin{itemize}
    \item ImageNet1k - classification: \\ \textcolor{confBlue}{microsoft/swin-tiny-patch4-window7-224}
    \item EuroSat - RS classification: \\ \textcolor{confBlue}{nielsr/swin-tiny-patch4-window7-224-finetuned-eurosat}
    \item ADE20k - semantic segmentation: \\ \textcolor{confBlue}{facebook/mask2former-swin-tiny-ade-semantic}
    \item COCO - panoptic segmentation: \\ \textcolor{confBlue}{facebook/mask2former-swin-tiny-coco-panoptic}
    \item COCO - instance segmentation: \\ \textcolor{confBlue}{facebook/mask2former-swin-tiny-coco-instance}
    \item CityScapes - panoptic segmentation: \\ \textcolor{confBlue}{facebook/mask2former-swin-tiny-cityscapes-panoptic}
    \item CityScapes - instance segmentation: \\ \textcolor{confBlue}{facebook/mask2former-swin-tiny-cityscapes-instance}
    \item CityScapes - semantic segmentation: \\ \textcolor{confBlue}{facebook/mask2former-swin-tiny-cityscapes-semantic}
    \item CityScapes - semantic segmentation (SwinB): \\ \textcolor{confBlue}{facebook/mask2former-swin-base-IN21k-cityscapes-semantic}
\end{itemize}

The hyperparameters all stays the same as explained in \Cref{subsec:impl_det}.

\subsubsection{Pre-training Task}
\label{asub:pt_task}

We report only the best performing dataset-task pairs in \Cref{subsec:pt-data} of main paper. Here, we further analyize different segmentation tasks -- semantic, instance, and panoptic -- available for CityScapes~\cite{cordts2016cityscapes} and COCO~\cite{lin2014coco}. The results are summarized in Table~\ref{tab:task}.

\begin{table}[!h]
    \caption{Comparison of different pre-training dataset-task combinations used during pre-training of the backbone. \textit{Pan} stands for panoptic, \textit{Sem} for semantic, and \textit{Inst} for instance segmentation.}
\resizebox{\linewidth}{!}{
    \setlength{\tabcolsep}{2pt}
    \centering
    \begin{tabular}{lccccccc}
    \toprule
    	~& SYSU& LEVIR& EGYBCD& GVLM& CLCD& OSCD& \textit{Avg}\\ \midrule
		 City-Sem & $\mathbf{82.1}$ & $\mathbf{91.0}$ & $83.1$ & $\mathbf{88.9}$ & $\mathbf{75.3}$ & $\mathbf{52.4}$ & $\mathbf{78.8}$\\
		 City-Pan & $81.5$ & $90.9$ & $83.5$ & $88.5$ & $75.2$ & $51.0$ & $78.4$\\
		 City-Inst & $81.6$ & $90.9$ & $\mathbf{83.6}$ & $88.8$ & $73.4$ & $46.8$ & $77.5$\\
		 COCO-Pan & $81.7$ & $90.8$ & $83.3$ & $88.5$ & $74.6$ & $49.8$ & $78.1$\\
		 COCO-Inst & $81.6$ & $\mathbf{91.0}$ & $\mathbf{83.6}$ & $88.8$ & $73.9$ & $43.8$ & $77.1$\\

    \bottomrule
    \end{tabular}
    }
    \label{tab:task}
\end{table}
It can be observed that panoptic and semantic segmentation consistently outperform instance segmentation, regardless of the pre-training dataset. The only exception is EGYBCD datasets, where instance segmentation yields the best performance.

\subsection{Backbone Architecture}
\label{asub:arch}

In this subsection, we provide implementation details for different architectures used, along with additional information such as parameter count and computational efficiency metrics.

We use architecture implementations from Huggingface\footnote{\url{https://huggingface.co/models}} in our experiments. All models are pretrained on ImageNet1k~\cite{deng2009imagenet} classification and are as follows:
\begin{itemize}
    \item ResNet18: \textcolor{confBlue}{microsoft/resnet-18}
    \item ResNet50: \textcolor{confBlue}{microsoft/resnet-50}
    \item ConvNextB: \textcolor{confBlue}{facebook/convnext-base-224}
    \item ViT-T: \textcolor{confBlue}{vit\_tiny\_patch16\_224}
    \item ViT-B: \textcolor{confBlue}{vit\_base\_patch16\_224}
    \item SwinV2-T: \textcolor{confBlue}{microsoft/swinv2-tiny-patch4-window8-256}
    \item Swin-T: \textcolor{confBlue}{microsoft/swin-tiny-patch4-window7-224}
\end{itemize}

All architectures follow the same set of hyperparameters as introduced in \Cref{subsec:impl_det}, except ViT architectures that use base learning rate of $6\cdot10^{-5}$ and weight decay of $0.05$, which works better in this case.
We also report parameter count, computational efficiency and change detection performance in \Cref{tab:arch_params}.

\begin{table}[!h]
    \caption{Parameter count (encoder, decoder, total), efficiency metrics (GFLOPs and FPS), and change detection performance across 6 datasets (F1) for different architectures.}
\resizebox{\linewidth}{!}{
    \setlength{\tabcolsep}{4pt}
    \centering
        \begin{tabular}{lcccccc}
        \toprule
 & \makecell{Enc.\\Param.} & \makecell{Dec.\\Param.} & \makecell{Total\\Param.} & GFLOPs & \makecell{FPS\\\relax[img/s]} & F1 \\
\midrule
ResNet18 & 11.2 & 29.6 & 40.8 & 114.2 & 180.4 & 75.3 \\
ResNet50 & 23.5 & 40.5 & 64.0 & 128.4 & 100.4 & 75.6 \\
ConvNext-B & 87.6 & 33.2 & 120.8 & 185.7 & 72.5 & 76.3 \\
ViT-T & 5.5 & 27.5 & 33.1 & 19.5 & 116.8 & 75.0 \\
ViT-B & 85.8 & 32.3 & 118.1 & 103.7 & 118.2 & 75.4 \\
SwinV2-T & 27.6 & 31.4 & 59.0 & 128.8 & 47.6 & 77.5 \\
Swin-T & 27.5 & 31.4 & 58.9 & 134.5 & 57.8 & 77.9 \\
        \bottomrule
        \end{tabular}
        }

    \label{tab:arch_params}
\end{table}

\subsection{Backbone size}
\label{asub:bb_size}

\Cref{tab:size_params} reports parameter count, computational efficiency and change detection performance for different sizes of Swin backbone: tiny, small and base.

\begin{table}[!h]
    \caption{Parameter count (encoder, decoder, total), efficiency metrics (GFLOPs and FPS), and change detection performance (F1) for different Swin backbone sizes.}
\resizebox{\linewidth}{!}{
    \setlength{\tabcolsep}{4pt}
    \centering
        \begin{tabular}{lcccccc}
        \toprule
 & \makecell{Enc.\\Param.} & \makecell{Dec.\\Param.} & \makecell{Total\\Param.} & GFLOPs & \makecell{FPS\\\relax[img/s]} & F1 \\
\midrule
swin-T & 27.5 & 31.4 & 58.9 & 134.5 & 57.8 & 77.9 \\
swin-S & 48.8 & 31.4 & 80.3 & 162.3 & 32.2 & 77.8 \\
swin-B & 86.9 & 33.2 & 120.1 & 221.4 & 32.1 & 78.4 \\
        \bottomrule
        \end{tabular}
        }

    \label{tab:size_params}
\end{table}

\subsection{Augmentations}
\label{asub:aug_imp}

This subsection provides implementation details for the augmentations used in our study. We use Albumentations\footnote{\url{https://albumentations.ai/}} library to perform the same augmentation on a pair of images and their label.

Before augmentation, images are always normalised with ImageNet1k mean and std and, if needed, resized to $256\times256$ pixels. We don't use any other random augmentations during testing.

For experiments with random augmentations during training, we use the following configuration:
For \textit{Flip} augmentation, horizontal flip, vertical flip, and random rotation on interval $[-90, 90]$ are used. For \textit{Crop}, a random crop with a ratio on interval $[0.3, 1]$ is performed. In case of \textit{Color} augmentation, brightness, contrast and saturation are scaled by a randomly selected factor on interval $[0.7, 1.3]$, while the factor for hue lies on the interval from $-0.05$ to $0.05$. In \textit{Blur} augmentation, the size of the Gaussian kernel is randomly selected from interval $[3, 9]$ and the sigma is computed to satisfy: $kernel size=\text{int}(sigma \cdot 3.5) \cdot 2 + 1$. 
All augmentations are applied with a $30~\%$ probability, and the same augmentation is always applied to both images in a pair.

\subsection{Learning rate schedulers}
\label{asub:sched}

Following the related works~\cite{daudt2018fcn, wang2024mtp, zheng2023changen}, we use the following hyperparameters for learning rate schedulers. Cosine scheduler uses no restarts. Exponential scheduler uses gamma of 0.95, while polynomial uses 0.9. Multistep scheduler multiplies the learning rate by a factor of 0.5 at $80~\%$ and $90~\%$ of epochs.

\subsection{Related Methods' Implementation Details}
\label{asub:relat_imp}

In this subsection, we provide implementation details for all related methods used in our paper. The details for remote sensing foundation models are in \Cref{asub:rspt_impl} and for change detection specific approaches in \Cref{asub:cd_impl}. The evaluation framework is the same for all methods and is explained in \Cref{subsec:eval} of the main paper

\subsubsection{Remote Sensing Foundation Models}
\label{asub:rspt_impl}

We use the official code and pre-trained weights provided by authors for all remote sensing foundation models. The specific versions of the code used in our experiments can be accessed through the following GitHub links:

\begin{itemize}
    \item SeCo~\cite{manas2021seco}: \url{https://github.com/ServiceNow/seasonal-contrast/commit/8285173ec205b64bc3e53b880344dd6c3f79fa7a}
    \item CaCo~\cite{mall2023caco}: \url{https://github.com/utkarshmall13/CACo/commit/7c4fbc8d9e65a4ef715596c5c275ab3ddb6a8193}
    \item GaSSL~\cite{ayush2021gessl}: \url{https://github.com/sustainlab-group/geography-aware-ssl/commit/425173156d33a9e26ec1682e4ddee6712406308c}
    \item SatMAE~\cite{cong2022satmae}: \url{https://github.com/sustainlab-group/SatMAE/commit/e31c11fa1bef6f9a9aa3eb49e8637c8b8952ba5e}
    \item GFM~\cite{mendieta2023gfm}: \url{https://github.com/mmendiet/GFM/commit/4dd248e8544b3b6a49f5173b0931d97a17a7f424}
    \item MTP~\cite{wang2024mtp}: \url{https://github.com/ViTAE-Transformer/MTP/commit/962f7fd8781c095eb26db65ead3016e666b6d417}
\end{itemize}

We adopt the authors' architecture code and load the weights as the backbone into our framework. The configuration is the same as for our best model, with minor adjustments according to authors' settings.

All methods have an initial learning rate of $10^{-4}$ (same as our model), except ViT based models MTP and SatMAE, to $6\cdot10^{-5}$ as per MTP~\cite{wang2024mtp}. All methods have weight decay set to $0.05$, except GFM which uses $5e-4$~\cite{mendieta2023gfm}.
We extract features from all four hierarchical levels for SeCo, CaCo, GaSSL and GFM. Following the practice in MTP~\cite{wang2024mtp} layers with ids 3, 5, 7, and 11 are used for MTP and for SatMAE layers with ids 7, 11, 15, and 23. All methods employ the UPerNet~\cite{xiao2018upernet} decoder, except the ViT based methods, where UNet~\cite{ronneberger2015u} decoder performs better and also aligns with the MTP configuration~\cite{wang2024mtp}. Other settings, including dice loss, cosine scheduler, and flip augmentations, remain the same as for our best model.

\subsubsection{Change Detection Specific Methods}
\label{asub:cd_impl}

We use official code, provided by authors, for all methods and just integrate our datasets. The following are links to used versions of their code on GitHub:

\begin{itemize}
    \item FCS-Diff~\cite{daudt2018fcn}:  \url{https://github.com/rcdaudt/fully_convolutional_change_detection/commit/4dd83231f25319a7ebb16cbfa9912541ceabac9a}
    \item BIT~\cite{chen2021bit}:  \url{https://github.com/justchenhao/BIT_CD/commit/adcd7aea6f234586ffffdd4e9959404f96271711}
    \item ChangeFormer~\cite{bandara2022changeFormer}:  \url{https://github.com/wgcban/ChangeFormer/commit/afd1b7ed640aa265a2c730de958416ae7356a2f9}
    \item BiFA~\cite{zhang2024bifa}:  \url{https://github.com/zmoka-zht/BiFA/commit/56cd0da461e5e4b0d6a9b4f3321f0a81a91d21b8}
    \item SwinSUNet~\cite{zhang2022swinsunet}:  \url{https://github.com/GuitarZhang/SwinSUNet/commit/721daf84238eda40fb49d626c21df4ed2246aa9e}
\end{itemize}

We keep all the hyperparameters same as set by authors, except epoch number for BiFA and FC-Siam-Diff. For these two models, we halve the default number of epochs in case of OSCD~\cite{daudt2018urban} dataset, since they start to overfit due to the dataset's small size.

\subsection{Transfer To Other Methods - Full Results}
\label{asub:trans_ext}

The full results of the experiment in \Cref{sec:transfer} are presented here.
We apply our insights to related foundation models (SeCo, CaCo, SatMAE, GFM, GaSSL and MTP) and change detection specific architectures FC-Siam-Diff~\cite{daudt2018fcn} and SwinSUNet~\cite{zhang2022swinsunet}. 

Performance is initially assessed under the base configuration: no augmentations, no scheduler, CE loss for \textit{foundation models}, and author default configuration for \textit{change detection specific models}. We then incorporate flip augmentations, dice loss, and a cosine scheduler. For FC-Siam-Diff and SwinSUNet, we also replace their default encoder with Swin-B pre-trained on the Cityscapes, which is also used in our approach (details in \Cref{asub:pt_impl}). 

The results in \Cref{atab:ext_trans} show consistent improvement of few percentage points across all models. Improvement is especially pronounced in case of FC-Siam-Diff~\cite{daudt2018fcn}, likely due to the fact that we introduce a powerful pretrained backbone, compared to its original non-pre-trained architecture.

\begin{table}[!ht]
    \caption{Average F1 across 6 datasets for different SOTA related methods with and without using our optimisations.}

\resizebox{\linewidth}{!}{
    \setlength{\tabcolsep}{3pt}
    \centering
    \begin{tabular}{lccccccc}
    \toprule
    	~& SYSU& LEVIR& EGYBCD& GVLM& CLCD& OSCD& \textit{Avg}\\ \midrule
		 SeCo~\cite{manas2021seco} & $79.1$ & $89.9$ & $82.5$ & $88.4$ & $65.4$ & $38.5$ & $74.0$\\
		 \rowcolor{opti!15}+ our opti. & $78.9$ & $90.6$ & $84.1$ & $90.4$ & $68.7$ & $49.1$& $77.0$ (\inc{3.0})\\
		 CaCo~\cite{mall2023caco} & $80.3$ & $90.1$ & $82.7$ & $89.3$ & $71.9$ & $40.2$ & $75.7$\\
		 \rowcolor{opti!15}+ our opti. & $80.2$ & $90.9$ & $84.8$ & $90.3$ & $77.9$ & $51.9$& $79.3$ (\inc{3.6})\\
		 SatMAE~\cite{cong2022satmae} & $78.9$ & $90.8$ & $84.2$ & $88.3$ & $75.9$ & $44.7$ & $77.1$\\
		 \rowcolor{opti!15}+ our opti. & $81.4$ & $91.4$ & $85.9$ & $89.8$ & $79.2$ & $49.5$& $79.5$ (\inc{2.4})\\
		 GFM~\cite{mendieta2023gfm} & $81.4$ & $89.1$ & $81.5$ & $88.7$ & $75.6$ & $50.1$ & $77.7$\\
		 \rowcolor{opti!15}+ our opti. & $81.2$ & $89.8$ & $85.0$ & $89.8$ & $77.5$ & $54.1$& $79.6$ (\inc{1.9})\\
		 GaSSL~\cite{ayush2021gessl} & $80.4$ & $90.3$ & $83.2$ & $89.2$ & $72.7$ & $39.4$ & $75.9$\\
		 \rowcolor{opti!15}+ our opti. & $80.2$ & $91.2$ & $85.4$ & $90.2$ & $78.1$ & $53.1$& $79.7$ (\inc{3.8})\\
		 MTP~\cite{wang2024mtp} & $82.1$ & $91.1$ & $84.6$ & $88.9$ & $74.4$ & $42.2$ & $77.2$\\
		 \rowcolor{opti!15}+ our opti. & $81.3$ & $91.7$ & $85.9$ & $89.9$ & $80.3$ & $52.8$& $80.3$ (\inc{3.1})\\
		 FCS-Diff\cite{daudt2018fcn} & $70.8$ & $81.8$ & $42.3$ & $74.3$ & $54.1$ & $39.4$ & $60.5$\\
		 \rowcolor{opti!15}+ our opti. & $79.3$ & $86.9$ & $80.4$ & $88.6$ & $73.2$ & $56.1$& $77.4$ (\inc{16.9})\\
		 SwinSUNet~\cite{zhang2022swinsunet} & $76.6$ & $89.3$ & $83.7$ & $90.0$ & $75.8$ & $52.8$ & $78.0$\\
		 \rowcolor{opti!15}+ our opti. & $81.8$ & $90.1$ & $84.6$ & $90.6$ & $80.3$ & $50.1$& $79.6$ (\inc{1.6})\\

    \bottomrule
    \end{tabular}
    }
    \label{atab:ext_trans}
\end{table}

\subsection{Computational Efficiency}
\label{sub:comp}

\begin{table*}[!t]
    \caption{Extended computational efficiency results for each model. We report backbone architecture, pretrain dataset, parameter count, inference time, FPS (derived from inference time), GFLOPs and average F1 across 6 datasets. All results were obtained using an Nvidia A100-SXM4 40GB GPU.}
    \centering
        \begin{tabular}{lccccccc}
        \toprule
 & Backbone & Pre-train & \makecell{Param. [M]} & \makecell{Inference\\time [ms]} & \makecell{FPS\\\relax[img/s]} & GFLOPs & F1 \\
\midrule
\cellcolor{orange!15}FCS-Diff~\cite{daudt2018fcn} & UNet & - & 1.4 & $5.9 \pm 0.0$ & $170.1 \pm 0.02$ & 4.6 & 60.5 \\
\cellcolor{orange!15}BIT~\cite{chen2021bit} & RN-18 & IN1k & 12.4 & $17.3 \pm 0.06$ & $57.7 \pm 0.19$ & 21.6 & 73.0 \\
\cellcolor{orange!15}ChangeFormer~\cite{bandara2022changeFormer} & MiT-B0* & ADE20k & 41.0 & $27.6 \pm 0.02$ & $36.2 \pm 0.03$ & 234.6 & 73.8 \\
\cellcolor{orange!15}BiFA~\cite{zhang2024bifa} & MiT-B0 & ADE20k & 9.9 & $31.0 \pm 0.04$ & $32.2 \pm 0.04$ & 4.3 & 76.3 \\
\cellcolor{TealBlue!15}SeCo~\cite{manas2021seco} & RN-50 & RS & 64.0 & $9.9 \pm 0.02$ & $100.7 \pm 0.23$ & 128.5 & 77.0 \\
\cellcolor{orange!15}SwinSUNet~\cite{zhang2022swinsunet} & Swin-T & IN1k & 43.6 & $30.2 \pm 0.08$ & $33.1 \pm 0.09$ & 32.6 & 78.0 \\
\cellcolor{TealBlue!15}CaCo~\cite{mall2023caco} & RN-50 & RS & 64.0 & $9.9 \pm 0.02$ & $100.9 \pm 0.21$ & 128.5 & 79.3 \\
\cellcolor{TealBlue!15}SatMAE~\cite{cong2022satmae} & ViT-L & RS & 322.6 & $14.0 \pm 0.03$ & $71.2 \pm 0.14$ & 514.2 & 79.5 \\
\cellcolor{TealBlue!15}GFM~\cite{mendieta2023gfm} & Swin-B & RS & 120.5 & $22.3 \pm 0.05$ & $44.9 \pm 0.1$ & 109.2 & 79.6 \\
\cellcolor{TealBlue!15}GaSSL~\cite{ayush2021gessl} & RN-50 & RS & 64.0 & $10.0 \pm 0.07$ & $100.4 \pm 0.66$ & 128.5 & 79.7 \\
\cellcolor{TealBlue!15}MTP~\cite{wang2024mtp} & ViT-B + RVSA & RS & 107.8 & $32.1 \pm 0.02$ & $31.2 \pm 0.02$ & 196.9 & 80.3 \\
BTC {\footnotesize T} (Ours) & Swin-T & CityS. & 58.9 & $17.3 \pm 0.05$ & $57.8 \pm 0.17$ & 134.5 & 80.5 \\
\textbf{BTC {\footnotesize B}} (Ours) & Swin-B & CityS. & 120.1 & $30.8 \pm 0.03$ & $32.4 \pm 0.03$ & 221.4 & 80.9 \\
        \bottomrule
        \end{tabular}

    \label{atab:perf}
\end{table*}

Here we describe the implementation of our protocol for measuring computational efficiency (\Cref{asub:comp}) and provide some additional metrics like GFLOPs for our method and other state-of-the-art methods (\Cref{asub:comp_res}).

\subsubsection{Implementation Details}
\label{asub:comp}

We measure 3 different computational efficiency metrics: parameter count, inference time (also expressed as frames per second - FPS) and GFLOPs. GFLOPs are measured using the official PyTorch profiler\footnote{\url{https://pytorch.org/tutorials/recipes/recipes/profiler_recipe.html}}. 

For inference time measurement, we use a pair of RGB $256\times256$ float16 images and set models to float16 in all cases. We first perform 1000 forward passes as a warm-up. Then, we measure the time of the next 1000 forward passes and record the runtime. To ensure robustness, we repeat these steps five times and report the mean runtime of a single pass. All experiments were conducted on an Nvidia A100-SXM4 40GB GPU.

\subsubsection{Additional Results}
\label{asub:comp_res}

We report the extended results in \Cref{atab:perf}. The FPS is calculated from inference time with the equation: $FPS = 1~second / \text{inference time}$.

The results reveal that \textbf{GFLOPs} (billion floating-point operations per second) \textbf{do not correlate well with inference time}, despite being commonly used in related works. This is because modern GPUs execute compute-insensitive floating-point operations very efficiently due to high parallelization and dedicated hardware. However, more complex operations, typically performed at higher level, can cause significant slowdowns. Large foundation models, like SatMAE~\cite{cong2022satmae}, built with efficient vision transformers, achieve fast inference time of $14~ms$ despite having 322 million parameters and 512 GFLOPs. In contrast, BiFA~\cite{zhang2024bifa}, with only 9.9 million parameters and 4.3 GFLOPs, takes twice as long for inference due to computationally less efficient high-level operations (such as grid instantiations and sampling).
\begin{table}[!h]
    \caption{Encoder, decoder and total parameter count for remote-sensing foundation models.}

    \setlength{\tabcolsep}{4pt}
    \centering
    \begin{tabular}{lcccccc}
        \toprule
         & CaCo & SeCo & GaSSL & GFM & MTP & SatMAE \\
        \midrule
        \makecell{Enc.\\Param.} & 23.5 & 23.5 & 23.5 & 86.8 & 93.3 & 304.4 \\
        \makecell{Dec.\\Param.} & 40.5 & 40.5 & 40.5 & 33.7 & 14.4 & 18.3 \\
        \makecell{Total\\Param.} & 64.0 & 64.0 & 64.0 & 120.5 & 107.8 & 322.6 \\
        \bottomrule
    \end{tabular}
    \label{tab:sota_params}
\end{table}
In \Cref{tab:sota_params} we also report separate encoder and decoder parameters for foundation models. The encoder parameters directly correspond to the foundation model parameters since these models are used as the encoder (backbone) in our framework.

\section{Standard Deviation Results}
\label{a:std}

Standard deviation of our analysis is provided in \Cref{asub:std_analyisis} and also for state-of-the-art methods in \Cref{asub:sota_std}

\subsection{Standard Deviation of Analysis Results}
\label{asub:std_analyisis}

Results with standard deviation for pre-training dataset-task pairs are in \Cref{atab:task}, for different architectures in \Cref{atab:arch}, and for different Swin sizes in \Cref{atab:size}. Results with standard deviation for augmentation experiments are in \Cref{atab:aug}, for schedulers in \Cref{atab:sched}, and loss functions in \Cref{atab:loss}.

Results for experiment where we combine all our insights are in \Cref{atab:acum_std} and finally results for transfer to other methods in \Cref{atab:transfer_std}.

\begin{table}[!h]
    \caption{Comparison of different pre-training datasets for the Swin backbone. \textit{None} means that no pre-training dataset was used. \textit{Pan} stands for panoptic, \textit{Inst} stands for instance, and \textit{Sem} for semantic segmentation.}
\resizebox{\linewidth}{!}{
    \setlength{\tabcolsep}{2pt}
    \centering
    \begin{tabular}{lccccccc}
    \toprule
    	~& SYSU& LEVIR& EGY-BCD& GVLM& CLCD& OSCD& \textit{Avg}\\ \midrule
		 None & \meanwithstd{$77.0$}{0.3} & \meanwithstd{$88.2$}{0.1} & \meanwithstd{$76.7$}{0.3} & \meanwithstd{$87.5$}{0.4} & \meanwithstd{$62.7$}{2.2} & \meanwithstd{$37.0$}{12.0} & \meanwithstd{$71.5$}{1.7}\\ \midrule
		 IN1k & \meanwithstd{$82.0$}{0.2} & \meanwithstd{$\mathbf{91.0}$}{0.1} & \meanwithstd{$83.5$}{0.2} & \meanwithstd{$88.8$}{0.4} & \meanwithstd{$74.6$}{0.4} & \meanwithstd{$47.6$}{4.6} & \meanwithstd{$77.9$}{0.6}\\ \midrule
		 EuroSat & \meanwithstd{$81.5$}{0.2} & \meanwithstd{$\mathbf{91.0}$}{0.1} & \meanwithstd{$83.4$}{0.1} & \meanwithstd{$87.8$}{2.0} & \meanwithstd{$74.3$}{0.2} & \meanwithstd{$49.4$}{2.6} & \meanwithstd{$77.9$}{0.3}\\ \midrule
		 ADE20k & \meanwithstd{$81.5$}{0.2} & \meanwithstd{$\mathbf{91.0}$}{0.1} & \meanwithstd{$83.5$}{0.1} & \meanwithstd{$\mathbf{88.9}$}{0.1} & \meanwithstd{$74.5$}{0.7} & \meanwithstd{$50.9$}{0.9} & \meanwithstd{$78.4$}{0.0}\\ \midrule
		 City-Sem & \meanwithstd{$\mathbf{82.1}$}{0.5} & \meanwithstd{$\mathbf{91.0}$}{0.0} & \meanwithstd{$83.1$}{0.6} & \meanwithstd{$\mathbf{88.9}$}{0.1} & \meanwithstd{$\mathbf{75.3}$}{0.6} & \meanwithstd{$\mathbf{52.4}$}{1.8} & \meanwithstd{$\mathbf{78.8}$}{0.3}\\ \midrule
		 City-Pan & \meanwithstd{$81.5$}{0.3} & \meanwithstd{$90.9$}{0.1} & \meanwithstd{$83.5$}{0.3} & \meanwithstd{$88.5$}{0.8} & \meanwithstd{$75.2$}{0.9} & \meanwithstd{$51.0$}{0.9} & \meanwithstd{$78.4$}{0.4}\\ \midrule
		 City-Inst & \meanwithstd{$81.6$}{0.3} & \meanwithstd{$90.9$}{0.0} & \meanwithstd{$\mathbf{83.6}$}{0.3} & \meanwithstd{$88.8$}{0.1} & \meanwithstd{$73.4$}{0.7} & \meanwithstd{$46.8$}{3.7} & \meanwithstd{$77.5$}{0.6}\\ \midrule
		 COCO-Pan & \meanwithstd{$81.7$}{0.2} & \meanwithstd{$90.8$}{0.2} & \meanwithstd{$83.3$}{0.4} & \meanwithstd{$88.5$}{0.7} & \meanwithstd{$74.6$}{1.5} & \meanwithstd{$49.8$}{1.9} & \meanwithstd{$78.1$}{0.7}\\ \midrule
		 COCO-Inst & \meanwithstd{$81.6$}{0.5} & \meanwithstd{$\mathbf{91.0}$}{0.0} & \meanwithstd{$\mathbf{83.6}$}{0.3} & \meanwithstd{$88.8$}{0.1} & \meanwithstd{$73.9$}{1.1} & \meanwithstd{$43.8$}{2.2} & \meanwithstd{$77.1$}{0.3}\\

    \bottomrule
    \end{tabular}
    }

    \label{atab:task}
\end{table}

\begin{table}[!h]
    \caption{Comparison of different backbone architectures.}

\resizebox{\linewidth}{!}{
    \setlength{\tabcolsep}{2pt}
    \centering
    \begin{tabular}{lccccccc}
    \toprule
    	~& SYSU& LEVIR& EGY-BCD& GVLM& CLCD& OSCD& \textit{Avg}\\ \midrule
		 Swin T & \meanwithstd{$82.0$}{0.2} & \meanwithstd{$\mathbf{91.0}$}{0.1} & \meanwithstd{$83.5$}{0.2} & \meanwithstd{$88.8$}{0.4} & \meanwithstd{$74.6$}{0.4} & \meanwithstd{$\mathbf{47.6}$}{4.6} & \meanwithstd{$\mathbf{77.9}$}{0.6}\\ \midrule
		 SwinV2 T & \meanwithstd{$81.4$}{0.4} & \meanwithstd{$\mathbf{91.0}$}{0.1} & \meanwithstd{$\mathbf{84.1}$}{0.3} & \meanwithstd{$89.2$}{0.1} & \meanwithstd{$\mathbf{75.0}$}{1.2} & \meanwithstd{$44.5$}{3.3} & \meanwithstd{$77.5$}{0.7}\\ \midrule
		 ViT T & \meanwithstd{$\mathbf{82.5}$}{0.5} & \meanwithstd{$84.5$}{0.1} & \meanwithstd{$81.3$}{0.6} & \meanwithstd{$88.0$}{0.2} & \meanwithstd{$73.5$}{0.2} & \meanwithstd{$40.3$}{1.5} & \meanwithstd{$75.0$}{0.2}\\ \midrule
		 ViT B & \meanwithstd{$81.9$}{0.2} & \meanwithstd{$85.2$}{0.2} & \meanwithstd{$82.1$}{0.4} & \meanwithstd{$87.7$}{0.4} & \meanwithstd{$72.8$}{1.0} & \meanwithstd{$42.5$}{3.0} & \meanwithstd{$75.4$}{0.3}\\ \midrule
		 ResNet18 & \meanwithstd{$80.1$}{0.4} & \meanwithstd{$89.9$}{0.1} & \meanwithstd{$82.3$}{0.2} & \meanwithstd{$88.9$}{0.0} & \meanwithstd{$69.9$}{1.9} & \meanwithstd{$40.9$}{6.1} & \meanwithstd{$75.3$}{1.0}\\ \midrule
		 ResNet50 & \meanwithstd{$80.8$}{0.2} & \meanwithstd{$90.3$}{0.1} & \meanwithstd{$82.7$}{0.2} & \meanwithstd{$\mathbf{89.5}$}{0.2} & \meanwithstd{$72.5$}{0.8} & \meanwithstd{$37.8$}{6.2} & \meanwithstd{$75.6$}{0.9}\\ \midrule
		 ConvNext B & \meanwithstd{$81.5$}{0.1} & \meanwithstd{$90.7$}{1.1} & \meanwithstd{$83.7$}{0.2} & \meanwithstd{$88.8$}{0.9} & \meanwithstd{$73.0$}{2.1} & \meanwithstd{$40.2$}{5.9} & \meanwithstd{$76.3$}{1.2}\\

    \bottomrule
    \end{tabular}
    }
    \label{atab:arch}
\end{table}

\begin{table}[!h]
    \caption{Comparison of different Swin backbone sizes: Tiny, Small and Base.}

\resizebox{\linewidth}{!}{
    \setlength{\tabcolsep}{2pt}
    \centering
    \begin{tabular}{lccccccc}
    \toprule
    	~& SYSU& LEVIR& EGY-BCD& GVLM& CLCD& OSCD& \textit{Avg}\\ \midrule
		 Swin T & \meanwithstd{$\mathbf{82.0}$}{0.2} & \meanwithstd{$91.0$}{0.1} & \meanwithstd{$83.5$}{0.2} & \meanwithstd{$88.8$}{0.4} & \meanwithstd{$74.6$}{0.4} & \meanwithstd{$\mathbf{47.6}$}{4.6} & \meanwithstd{$77.9$}{0.6}\\ \midrule
		 Swin S & \meanwithstd{$81.8$}{0.5} & \meanwithstd{$91.1$}{0.0} & \meanwithstd{$84.1$}{0.1} & \meanwithstd{$89.2$}{0.2} & \meanwithstd{$75.8$}{0.1} & \meanwithstd{$45.0$}{3.5} & \meanwithstd{$77.8$}{0.6}\\ \midrule
		 Swin B & \meanwithstd{$\mathbf{82.0}$}{0.5} & \meanwithstd{$\mathbf{91.2}$}{0.3} & \meanwithstd{$\mathbf{84.4}$}{0.1} & \meanwithstd{$\mathbf{89.4}$}{0.2} & \meanwithstd{$\mathbf{76.3}$}{1.6} & \meanwithstd{$47.3$}{5.3} & \meanwithstd{$\mathbf{78.4}$}{1.1}\\ 

    \bottomrule
    \end{tabular}
    }
    \label{atab:size}
\end{table}

\begin{table}[!h]
    \caption{Comparison of different augmentation techniques used during training. \textit{Flip} and \textit{Crop} serve as data expanders, while \textit{Color} and \textit{Blur} can be used as countermeasures for photographic variations in data.}
\resizebox{\linewidth}{!}{
    \setlength{\tabcolsep}{2pt}
    \centering
    \begin{tabular}{lccccccc}
    \toprule
    	~& SYSU& LEVIR& EGY-BCD& GVLM& CLCD& OSCD& \textit{Avg}\\ \midrule
		 None & \meanwithstd{$\mathbf{82.0}$}{0.2} & \meanwithstd{$91.0$}{0.1} & \meanwithstd{$83.5$}{0.2} & \meanwithstd{$88.8$}{0.4} & \meanwithstd{$74.6$}{0.4} & \meanwithstd{$47.6$}{4.6} & \meanwithstd{$77.9$}{0.6}\\ \midrule
		 Flip & \meanwithstd{$81.3$}{0.4} & \meanwithstd{$\mathbf{91.4}$}{0.0} & \meanwithstd{$\mathbf{85.4}$}{0.3} & \meanwithstd{$\mathbf{90.0}$}{0.2} & \meanwithstd{$78.7$}{0.4} & \meanwithstd{$\mathbf{51.0}$}{1.5} & \meanwithstd{$\mathbf{79.6}$}{0.3}\\ \midrule
		 Crop & \meanwithstd{$81.8$}{0.5} & \meanwithstd{$91.1$}{0.3} & \meanwithstd{$85.2$}{0.2} & \meanwithstd{$89.3$}{0.5} & \meanwithstd{$77.4$}{0.5} & \meanwithstd{$50.9$}{2.0} & \meanwithstd{$79.3$}{0.3}\\ \midrule
		 Flip \& Crop & \meanwithstd{$80.3$}{0.5} & \meanwithstd{$\mathbf{91.4}$}{0.1} & \meanwithstd{$84.8$}{1.3} & \meanwithstd{$\mathbf{90.0}$}{0.4} & \meanwithstd{$\mathbf{79.3}$}{0.4} & \meanwithstd{$49.1$}{4.4} & \meanwithstd{$79.2$}{0.8}\\ \midrule
		 Color & \meanwithstd{$81.7$}{0.5} & \meanwithstd{$91.0$}{0.1} & \meanwithstd{$83.7$}{0.0} & \meanwithstd{$89.0$}{0.1} & \meanwithstd{$74.4$}{0.2} & \meanwithstd{$47.9$}{3.5} & \meanwithstd{$77.9$}{0.6}\\ \midrule
		 Blur & \meanwithstd{$81.4$}{0.4} & \meanwithstd{$91.1$}{0.2} & \meanwithstd{$83.6$}{0.4} & \meanwithstd{$88.9$}{0.4} & \meanwithstd{$74.8$}{0.8} & \meanwithstd{$46.5$}{5.4} & \meanwithstd{$77.7$}{0.8}\\

    \bottomrule
    \end{tabular}
    }

    \label{atab:aug}
\end{table}

\begin{table}[!h]
    \caption{Comparison of different learning rate schedulers.}

\resizebox{\linewidth}{!}{
    \setlength{\tabcolsep}{2pt}
    \centering
    \begin{tabular}{lccccccc}
    \toprule
    	~& SYSU& LEVIR& EGY-BCD& GVLM& CLCD& OSCD& \textit{Avg}\\ \midrule
		 None & \meanwithstd{$82.0$}{0.2} & \meanwithstd{$\mathbf{91.0}$}{0.1} & \meanwithstd{$\mathbf{83.5}$}{0.2} & \meanwithstd{$88.8$}{0.4} & \meanwithstd{$\mathbf{74.6}$}{0.4} & \meanwithstd{$\mathbf{47.6}$}{4.6} & \meanwithstd{$\mathbf{77.9}$}{0.6}\\ \midrule
		 Multistep & \meanwithstd{$81.9$}{0.1} & \meanwithstd{$90.8$}{0.1} & \meanwithstd{$82.8$}{0.5} & \meanwithstd{$\mathbf{88.9}$}{0.4} & \meanwithstd{$73.5$}{0.4} & \meanwithstd{$43.6$}{3.2} & \meanwithstd{$76.9$}{0.5}\\ \midrule
		 Cosine & \meanwithstd{$81.9$}{0.3} & \meanwithstd{$90.4$}{0.1} & \meanwithstd{$82.9$}{0.4} & \meanwithstd{$88.5$}{0.1} & \meanwithstd{$73.5$}{0.4} & \meanwithstd{$45.3$}{0.8} & \meanwithstd{$77.1$}{0.2}\\ \midrule
		 Exp. & \meanwithstd{$\mathbf{82.2}$}{0.2} & \meanwithstd{$89.4$}{0.2} & \meanwithstd{$81.9$}{0.1} & \meanwithstd{$88.2$}{0.2} & \meanwithstd{$72.7$}{0.4} & \meanwithstd{$46.0$}{3.0} & \meanwithstd{$76.7$}{0.4}\\ \midrule
		 Lin. & \meanwithstd{$81.9$}{0.1} & \meanwithstd{$90.3$}{0.1} & \meanwithstd{$82.8$}{0.1} & \meanwithstd{$88.6$}{0.1} & \meanwithstd{$73.2$}{0.5} & \meanwithstd{$44.6$}{3.5} & \meanwithstd{$76.9$}{0.7}\\ \midrule
		 Poly. & \meanwithstd{$81.8$}{0.1} & \meanwithstd{$90.4$}{0.0} & \meanwithstd{$82.8$}{0.1} & \meanwithstd{$88.7$}{0.2} & \meanwithstd{$73.3$}{0.7} & \meanwithstd{$44.3$}{2.1} & \meanwithstd{$76.9$}{0.3}\\

    \bottomrule
    \end{tabular}
    }
    \label{atab:sched}
\end{table}

\begin{table}[!h]
    \caption{Comparison of different loss functions on change detection performance.}

\resizebox{\linewidth}{!}{
    \setlength{\tabcolsep}{2pt}
    \centering
    \begin{tabular}{lccccccc}
    \toprule
    	~& SYSU& LEVIR& EGY-BCD& GVLM& CLCD& OSCD& \textit{Avg}\\ \midrule
		 CE & \meanwithstd{$\mathbf{82.0}$}{0.2} & \meanwithstd{$91.0$}{0.1} & \meanwithstd{$83.5$}{0.2} & \meanwithstd{$88.8$}{0.4} & \meanwithstd{$74.6$}{0.4} & \meanwithstd{$47.6$}{4.6} & \meanwithstd{$77.9$}{0.6}\\ \midrule
		 Focal & \meanwithstd{$81.8$}{0.8} & \meanwithstd{$90.9$}{0.2} & \meanwithstd{$83.1$}{0.1} & \meanwithstd{$88.9$}{0.1} & \meanwithstd{$\mathbf{75.2}$}{0.8} & \meanwithstd{$44.1$}{6.6} & \meanwithstd{$77.3$}{1.2}\\ \midrule
		 Dice & \meanwithstd{$81.7$}{0.8} & \meanwithstd{$91.0$}{0.1} & \meanwithstd{$83.3$}{0.3} & \meanwithstd{$\mathbf{89.0}$}{0.2} & \meanwithstd{$74.7$}{0.1} & \meanwithstd{$\mathbf{53.0}$}{1.8} & \meanwithstd{$\mathbf{78.8}$}{0.4}\\ \midrule
		 Focal+Dice & \meanwithstd{$81.5$}{0.2} & \meanwithstd{$90.9$}{0.2} & \meanwithstd{$\mathbf{83.8}$}{0.1} & \meanwithstd{$\mathbf{89.0}$}{0.2} & \meanwithstd{$74.4$}{0.6} & \meanwithstd{$51.0$}{3.9} & \meanwithstd{$78.4$}{0.6}\\ \midrule
		 CE+Dice & \meanwithstd{$81.8$}{0.6} & \meanwithstd{$\mathbf{91.1}$}{0.1} & \meanwithstd{$83.7$}{0.2} & \meanwithstd{$\mathbf{89.0}$}{0.1} & \meanwithstd{$74.1$}{1.3} & \meanwithstd{$48.0$}{5.2} & \meanwithstd{$77.9$}{1.0}\\

    \bottomrule
    \end{tabular}
    }
    \label{atab:loss}
\end{table}

\begin{table}[!h]
    \caption{Results with included standard deviation when combining our insights to reach the final optimised change detection architecture. Each row presents the F1 score and the difference from the previous row. "+" indicates introduction of new design choices, while \switch{} represent exchange of a design choice with another.}
\resizebox{\linewidth}{!}{
    \setlength{\tabcolsep}{2pt}
    \centering
    \begin{tabular}{lccccccc}
    \toprule
    	~& SYSU& LEVIR& EGY-BCD& GVLM& CLCD& OSCD& \textit{Avg}\\ \midrule
		 Base & \meanwithstd{$77.0$}{0.3} & \meanwithstd{$88.2$}{0.1} & \meanwithstd{$76.7$}{0.3} & \meanwithstd{$87.5$}{0.4} & \meanwithstd{$62.7$}{2.2} & \meanwithstd{$37.0$}{12.0} & \meanwithstd{$71.5$}{1.7}\\ \midrule
		 + \makecell[l]{IN1k \\ pretrain} & \meanwithstd{$82.0$}{0.2} & \meanwithstd{$91.0$}{0.1} & \meanwithstd{$83.5$}{0.2} & \meanwithstd{$88.8$}{0.4} & \meanwithstd{$74.6$}{0.4} & \meanwithstd{$47.6$}{4.6} & \meanwithstd{$77.9$}{0.6}\\ \midrule
		 + \makecell[l]{Flip \\ augment} & \meanwithstd{$81.3$}{0.4} & \meanwithstd{$91.4$}{0.0} & \meanwithstd{$85.4$}{0.3} & \meanwithstd{$90.0$}{0.2} & \meanwithstd{$78.7$}{0.4} & \meanwithstd{$51.0$}{1.5} & \meanwithstd{$79.6$}{0.3}\\ \midrule
		 \switch{} \makecell[l]{CityS. \\ pretrain} & \meanwithstd{$81.0$}{0.1} & \meanwithstd{$91.5$}{0.2} & \meanwithstd{$85.9$}{0.2} & \meanwithstd{$90.1$}{0.1} & \meanwithstd{$79.1$}{0.7} & \meanwithstd{$54.8$}{3.0} & \meanwithstd{$80.4$}{0.4}\\ \midrule
		 + \makecell[l]{Cosine \\ scheduler} & \meanwithstd{$81.0$}{0.4} & \meanwithstd{$91.6$}{0.1} & \meanwithstd{$\mathbf{86.2}$}{0.1} & \meanwithstd{$90.5$}{0.1} & \meanwithstd{$79.4$}{0.9} & \meanwithstd{$54.9$}{0.6} & \meanwithstd{$80.6$}{0.2}\\ \midrule
		 \switch{} Swin-B & \meanwithstd{$81.8$}{0.2} & \meanwithstd{$\mathbf{91.7}$}{0.1} & \meanwithstd{$86.0$}{0.2} & \meanwithstd{$\mathbf{90.7}$}{0.2} & \meanwithstd{$\mathbf{81.6}$}{0.6} & \meanwithstd{$52.4$}{0.6} & \meanwithstd{$80.7$}{0.2}\\ \midrule
		 \switch{} Dice loss & \meanwithstd{$\mathbf{82.4}$}{0.4} & \meanwithstd{$91.5$}{0.1} & \meanwithstd{$85.6$}{0.1} & \meanwithstd{$\mathbf{90.7}$}{0.0} & \meanwithstd{$80.9$}{0.7} & \meanwithstd{$54.3$}{0.6} & \meanwithstd{$\mathbf{80.9}$}{0.0}\\

    \bottomrule
    \end{tabular}
    }

    \label{atab:acum_std}
\end{table}

\begin{table}[!h]
    \caption{Results with standard deviation for remote sensing foundation models without our insights and two change detection specific methods with our insights. For complementary results, refer to \Cref{atab:sota_std}}
\resizebox{\linewidth}{!}{
    \setlength{\tabcolsep}{2pt}
    \centering
    \begin{tabular}{lccccccc}
    \toprule
    	~& SYSU& LEVIR& EGY-BCD& GVLM& CLCD& OSCD& \textit{Avg}\\ \midrule
		 \makecell{SeCo~\cite{manas2021seco}\\ base} & \meanwithstd{$79.1$}{0.8} & \meanwithstd{$89.9$}{0.1} & \meanwithstd{$82.5$}{0.2} & \meanwithstd{$88.4$}{0.9} & \meanwithstd{$65.4$}{0.7} & \meanwithstd{$38.5$}{4.1} & \meanwithstd{$74.0$}{0.8}\\ \midrule
		 \makecell{CaCo~\cite{mall2023caco}\\ base} & \meanwithstd{$80.3$}{0.3} & \meanwithstd{$90.1$}{0.0} & \meanwithstd{$82.7$}{0.1} & \meanwithstd{$89.3$}{0.2} & \meanwithstd{$71.9$}{1.3} & \meanwithstd{$40.2$}{5.1} & \meanwithstd{$75.7$}{0.9}\\ \midrule
		 \makecell{GaSSL~\cite{ayush2021gessl}\\ base} & \meanwithstd{$80.4$}{0.7} & \meanwithstd{$90.3$}{0.0} & \meanwithstd{$83.2$}{0.1} & \meanwithstd{$89.2$}{0.1} & \meanwithstd{$72.7$}{1.5} & \meanwithstd{$39.4$}{8.4} & \meanwithstd{$75.9$}{1.7}\\ \midrule
		 \makecell{SatMAE~\cite{cong2022satmae}\\ base} & \meanwithstd{$78.9$}{1.0} & \meanwithstd{$90.8$}{0.1} & \meanwithstd{$84.2$}{0.3} & \meanwithstd{$88.3$}{1.0} & \meanwithstd{$75.9$}{0.8} & \meanwithstd{$44.7$}{3.1} & \meanwithstd{$77.1$}{0.5}\\ \midrule
		 \makecell{MTP~\cite{wang2024mtp} \\ base} & \meanwithstd{$\mathbf{82.1}$}{0.3} & \meanwithstd{$\mathbf{91.1}$}{0.1} & \meanwithstd{$\mathbf{84.6}$}{0.4} & \meanwithstd{$88.9$}{0.4} & \meanwithstd{$74.4$}{0.7} & \meanwithstd{$42.2$}{3.5} & \meanwithstd{$77.2$}{0.7}\\ \midrule
		 \makecell{GFM~\cite{mendieta2023gfm}\\ base} & \meanwithstd{$81.4$}{0.6} & \meanwithstd{$89.1$}{0.0} & \meanwithstd{$81.5$}{2.6} & \meanwithstd{$88.7$}{0.2} & \meanwithstd{$75.6$}{2.0} & \meanwithstd{$50.0$}{1.0} & \meanwithstd{$77.7$}{0.5}\\ \midrule
		 \makecell{FCS-Diff~\cite{daudt2018fcn}\\ +opti.} & \meanwithstd{$79.3$}{0.2} & \meanwithstd{$86.9$}{0.0} & \meanwithstd{$80.4$}{0.2} & \meanwithstd{$88.6$}{0.1} & \meanwithstd{$73.2$}{0.5} & \meanwithstd{$\mathbf{56.1}$}{0.7} & \meanwithstd{$77.4$}{0.1}\\ \midrule
		 \makecell{SwinSUNet~\cite{zhang2022swinsunet}\\ +opti} & \meanwithstd{$81.8$}{0.3} & \meanwithstd{$90.1$}{0.1} & \meanwithstd{$\mathbf{84.6}$}{0.2} & \meanwithstd{$\mathbf{90.6}$}{0.1} & \meanwithstd{$\mathbf{80.3}$}{0.3} & \meanwithstd{$50.1$}{1.3} & \meanwithstd{$\mathbf{79.6}$}{0.3}\\

    \bottomrule
    \end{tabular}
    }

    \label{atab:transfer_std}
\end{table}

\subsection{Standard Deviation of State-Of-The-Art Results}
\label{asub:sota_std}

The results with included standard deviation for our best approach and all other state-of-the-art methods are in \Cref{atab:sota_std}

\begin{table}[!h]
    \caption{Comparison of the proposed method to the state-of-the-art on six different datasets. We report the mean F1 of three runs with different random seeds for each dataset, and an average across all datasets. The standard deviation is also included for each result. \colorbox{TealBlue!15}{Remote sensing pre-training based} methods are highlighted with \colorbox{TealBlue!15}{blue color} and \colorbox{orange!15}{change detection specific} architectures are marked with \colorbox{orange!15}{orange}.}
\resizebox{\linewidth}{!}{
    \setlength{\tabcolsep}{2pt}
    \centering
    \begin{tabular}{lccccccc}
    \toprule
    	~& SYSU& LEVIR& EGY-BCD& GVLM& CLCD& OSCD& \textit{Avg}\\ \midrule
		 \cellcolor{orange!15}FCS-Diff~\cite{daudt2018fcn} & \meanwithstd{$70.8$}{0.1} & \meanwithstd{$81.8$}{0.6} & \meanwithstd{$42.3$}{4.0} & \meanwithstd{$74.3$}{0.7} & \meanwithstd{$54.1$}{1.1} & \meanwithstd{$39.4$}{1.3} & \meanwithstd{$60.5$}{0.5}\\
		 \cellcolor{orange!15}BIT~\cite{chen2021bit} & \meanwithstd{$77.0$}{0.1} & \meanwithstd{$90.0$}{0.1} & \meanwithstd{$75.9$}{1.5} & \meanwithstd{$88.8$}{0.1} & \meanwithstd{$63.4$}{2.0} & \meanwithstd{$42.6$}{2.0} & \meanwithstd{$73.0$}{0.4}\\
		 \cellcolor{orange!15}ChangeFormer~\cite{bandara2022changeFormer} & \meanwithstd{$77.9$}{0.5} & \meanwithstd{$89.5$}{0.2} & \meanwithstd{$77.9$}{0.6} & \meanwithstd{$88.5$}{0.2} & \meanwithstd{$60.8$}{2.6} & \meanwithstd{$48.1$}{1.0} & \meanwithstd{$73.8$}{0.3}\\
		 \cellcolor{orange!15}BiFA~\cite{zhang2024bifa} & \meanwithstd{$\mathbf{83.8}$}{0.3} & \meanwithstd{$89.5$}{0.9} & \meanwithstd{$83.5$}{0.2} & \meanwithstd{$89.0$}{0.2} & \meanwithstd{$74.5$}{0.6} & \meanwithstd{$37.4$}{3.7} & \meanwithstd{$76.3$}{0.8}\\
		 \cellcolor{TealBlue!15}SeCo~\cite{manas2021seco} & \meanwithstd{$78.9$}{0.1} & \meanwithstd{$90.6$}{0.1} & \meanwithstd{$84.1$}{0.1} & \meanwithstd{$90.4$}{0.1} & \meanwithstd{$68.7$}{0.9} & \meanwithstd{$49.1$}{1.4} & \meanwithstd{$77.0$}{0.3}\\
		 \cellcolor{orange!15}SwinSUNet~\cite{zhang2022swinsunet} & \meanwithstd{$76.6$}{1.5} & \meanwithstd{$89.3$}{0.1} & \meanwithstd{$83.7$}{0.0} & \meanwithstd{$90.0$}{0.4} & \meanwithstd{$75.8$}{0.7} & \meanwithstd{$52.8$}{2.6} & \meanwithstd{$78.0$}{0.4}\\
		 \cellcolor{TealBlue!15}CaCo~\cite{mall2023caco} & \meanwithstd{$80.2$}{0.4} & \meanwithstd{$90.9$}{0.0} & \meanwithstd{$84.8$}{0.1} & \meanwithstd{$90.3$}{0.0} & \meanwithstd{$77.9$}{0.5} & \meanwithstd{$51.9$}{1.0} & \meanwithstd{$79.3$}{0.0}\\
		 \cellcolor{TealBlue!15}SatMAE~\cite{cong2022satmae} & \meanwithstd{$81.4$}{0.1} & \meanwithstd{$91.4$}{0.1} & \meanwithstd{$\mathbf{85.9}$}{0.2} & \meanwithstd{$89.8$}{0.2} & \meanwithstd{$79.2$}{0.8} & \meanwithstd{$49.5$}{1.2} & \meanwithstd{$79.5$}{0.2}\\
		 \cellcolor{TealBlue!15}GFM~\cite{mendieta2023gfm} & \meanwithstd{$81.2$}{0.2} & \meanwithstd{$89.8$}{0.1} & \meanwithstd{$85.0$}{0.2} & \meanwithstd{$89.8$}{0.1} & \meanwithstd{$77.5$}{0.9} & \meanwithstd{$54.1$}{1.4} & \meanwithstd{$79.6$}{0.3}\\
		 \cellcolor{TealBlue!15}GaSSL~\cite{ayush2021gessl} & \meanwithstd{$80.2$}{0.3} & \meanwithstd{$91.2$}{0.1} & \meanwithstd{$85.4$}{0.2} & \meanwithstd{$90.2$}{0.1} & \meanwithstd{$78.1$}{0.8} & \meanwithstd{$53.1$}{1.0} & \meanwithstd{$79.7$}{0.2}\\
		 \cellcolor{TealBlue!15}MTP~\cite{wang2024mtp} & \meanwithstd{$81.3$}{0.3} & \meanwithstd{$\mathbf{91.7}$}{0.0} & \meanwithstd{$\mathbf{85.9}$}{0.3} & \meanwithstd{$89.9$}{0.2} & \meanwithstd{$80.3$}{0.3} & \meanwithstd{$52.8$}{0.4} & \meanwithstd{$80.3$}{0.0}\\
		 BTC {\footnotesize T} (Ours) & \meanwithstd{$81.6$}{0.1} & \meanwithstd{$91.4$}{0.1} & \meanwithstd{$85.6$}{0.1} & \meanwithstd{$90.6$}{0.1} & \meanwithstd{$79.5$}{0.5} & \meanwithstd{$\mathbf{54.3}$}{1.2} & \meanwithstd{$80.5$}{0.2}\\
		 \textbf{BTC {\footnotesize B}} (Ours) & \meanwithstd{$82.4$}{0.4} & \meanwithstd{$91.5$}{0.1} & \meanwithstd{$85.6$}{0.1} & \meanwithstd{$\mathbf{90.7}$}{0.0} & \meanwithstd{$\mathbf{80.9}$}{0.7} & \meanwithstd{$\mathbf{54.3}$}{0.6} & \meanwithstd{$\mathbf{80.9}$}{0.0}\\

    \bottomrule
    \end{tabular}
    }

    \label{atab:sota_std}
\end{table}

\section{Additional Qualitative Results}
\label{asub:qual}

More qualitative results are presented in this section. First, we provide qualitative results for all experiments of our analysis in \Cref{asub:ana_qual}. Then we also provide the qualitative results of all the state-of-the-art methods in \Cref{asub:sota_qual}.

\subsection{Analysis - Complete Qualitative Results}
\label{asub:ana_qual}

\begin{figure*}[!h]
    \centering
    \includegraphics[width=1\linewidth]{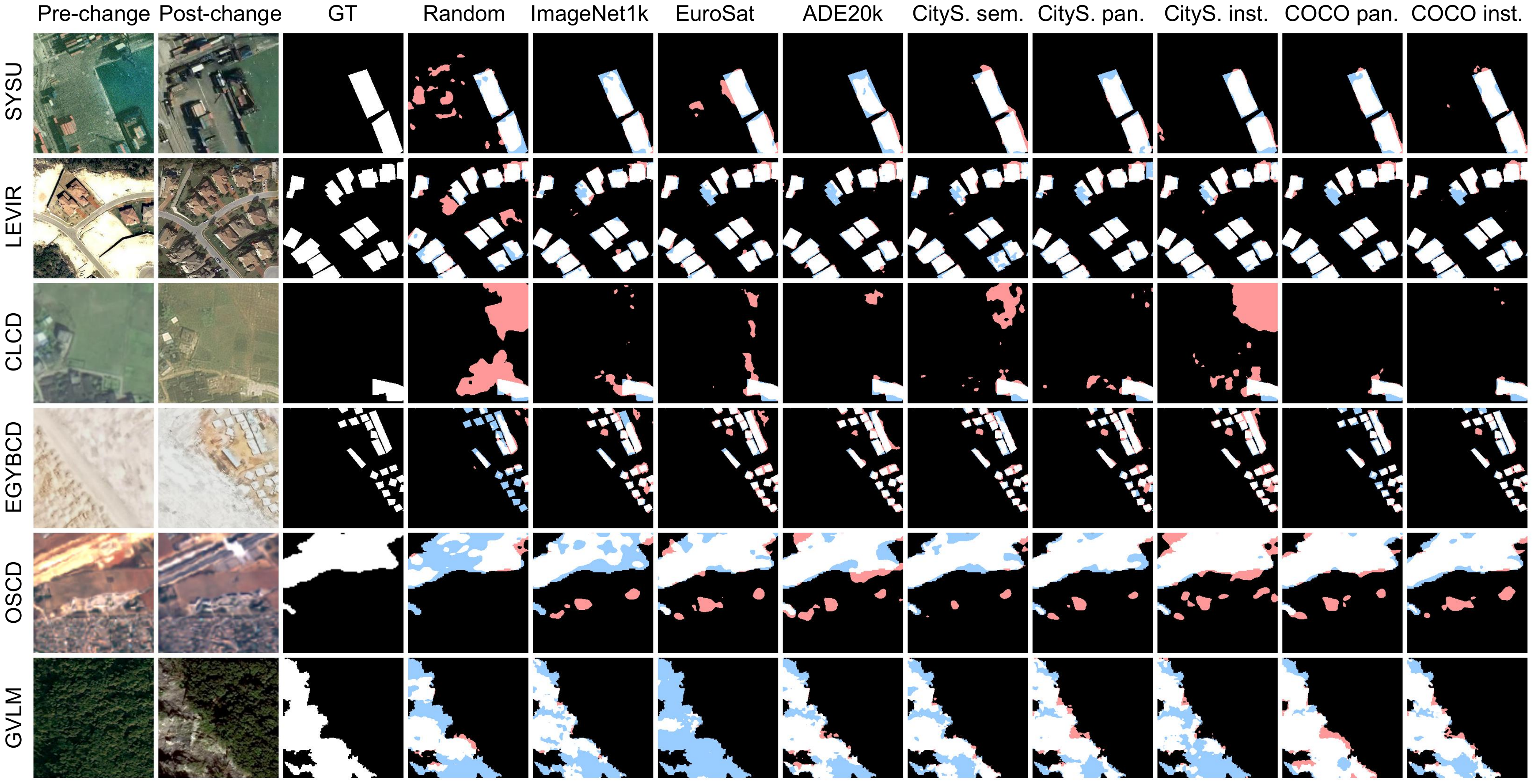}
    \caption{Additional qualitative examples for different backbone pre-training dataset-task pairs.}
    \label{afig:qual_pt}
\end{figure*}

\begin{figure*}[!h]
    \centering
    \includegraphics[width=1\linewidth]{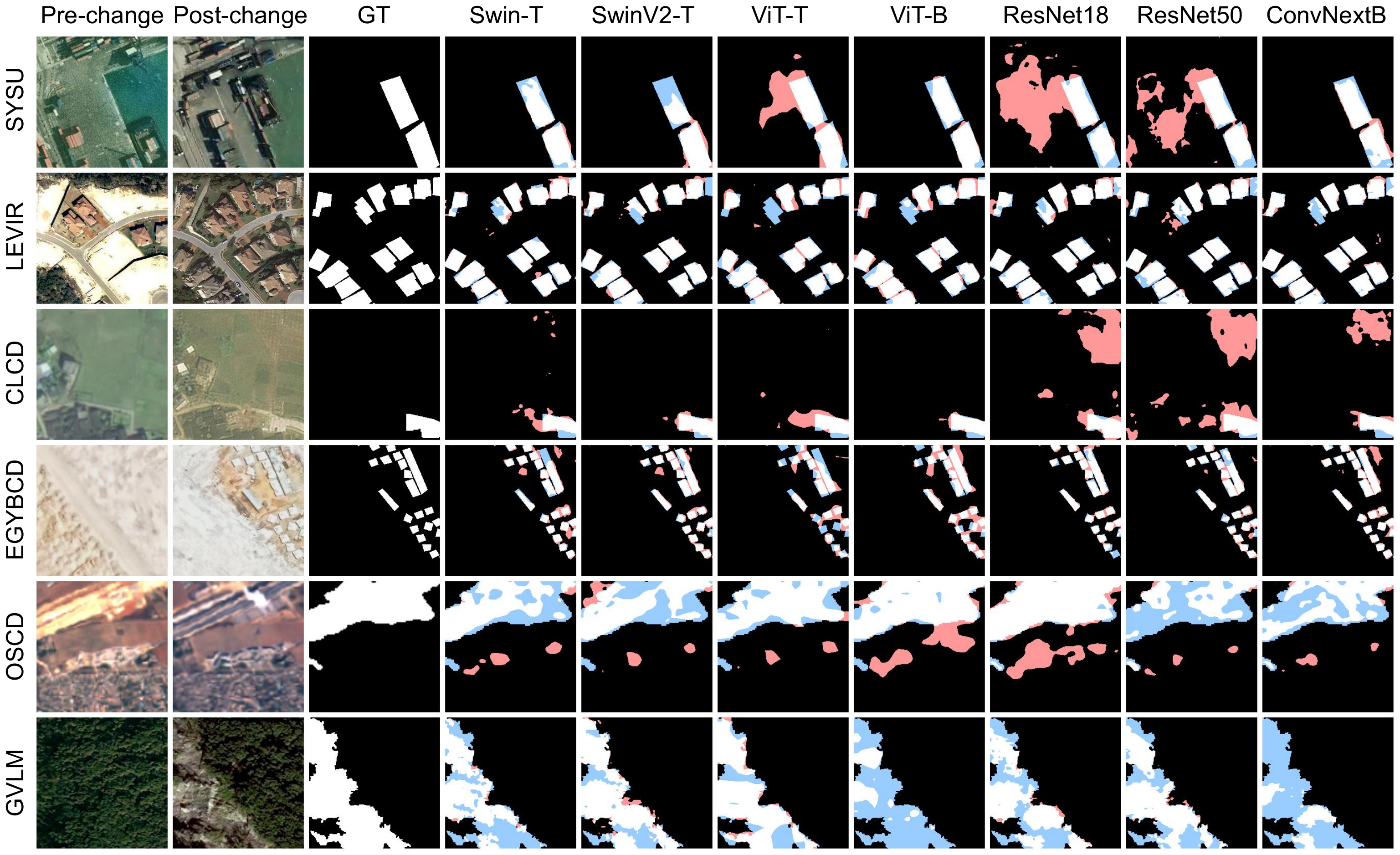}
    \caption{Additional qualitative examples for different architectures.}
    \label{afig:qual_arch}
\end{figure*}

\begin{figure*}[!h]
    \centering
    \includegraphics[width=0.6\linewidth]{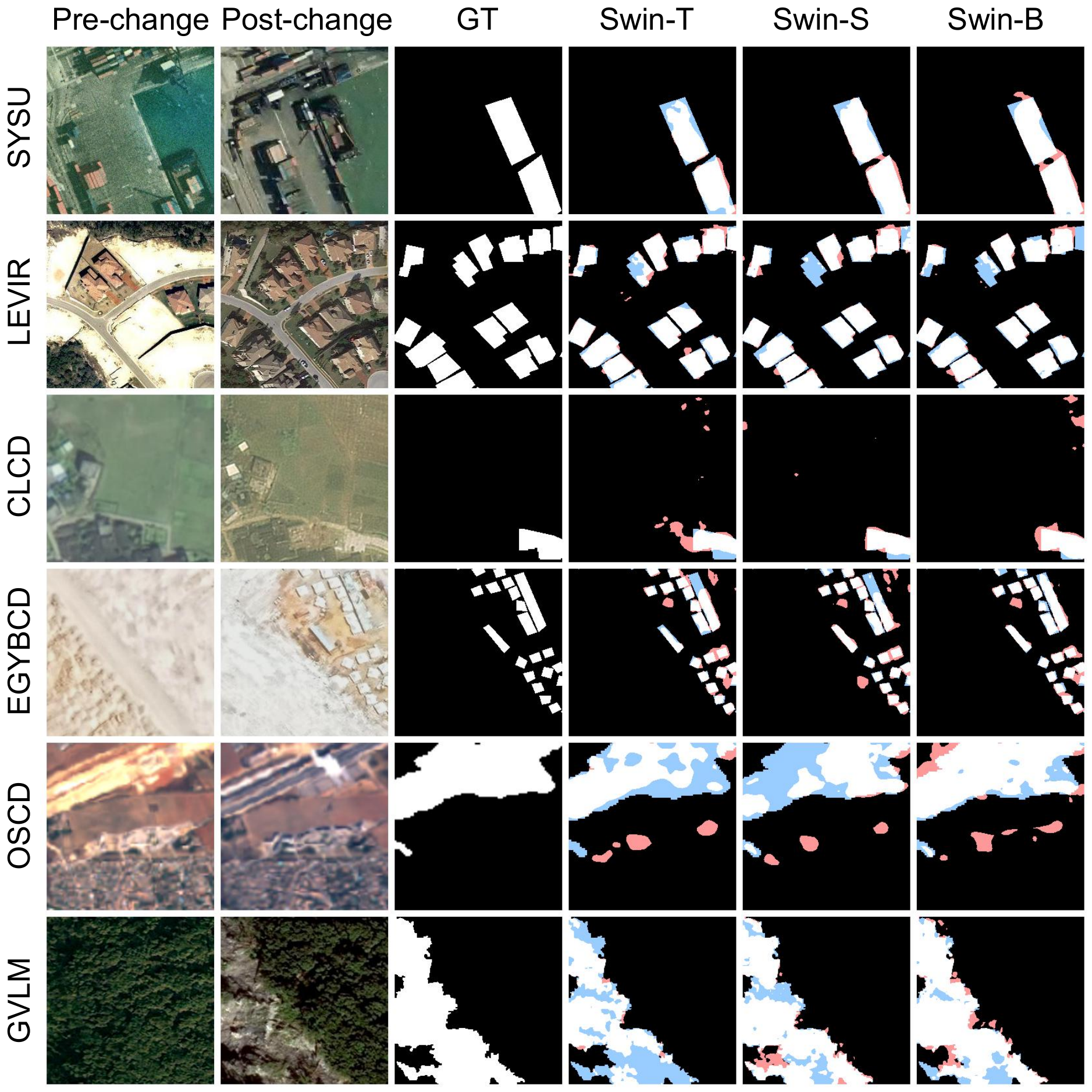}
    \caption{Additional qualitative examples for different Swin backbone sizes.}
    \label{afig:qual_sz}
\end{figure*}

Qualitative results for different backbone pre-training dataset-task pair are in \Cref{afig:qual_pt}, for different architectures in \Cref{afig:qual_arch}, and different Swin sizes in \Cref{afig:qual_sz}. 

\begin{figure*}[!h]
    \centering
    \includegraphics[width=0.8\linewidth]{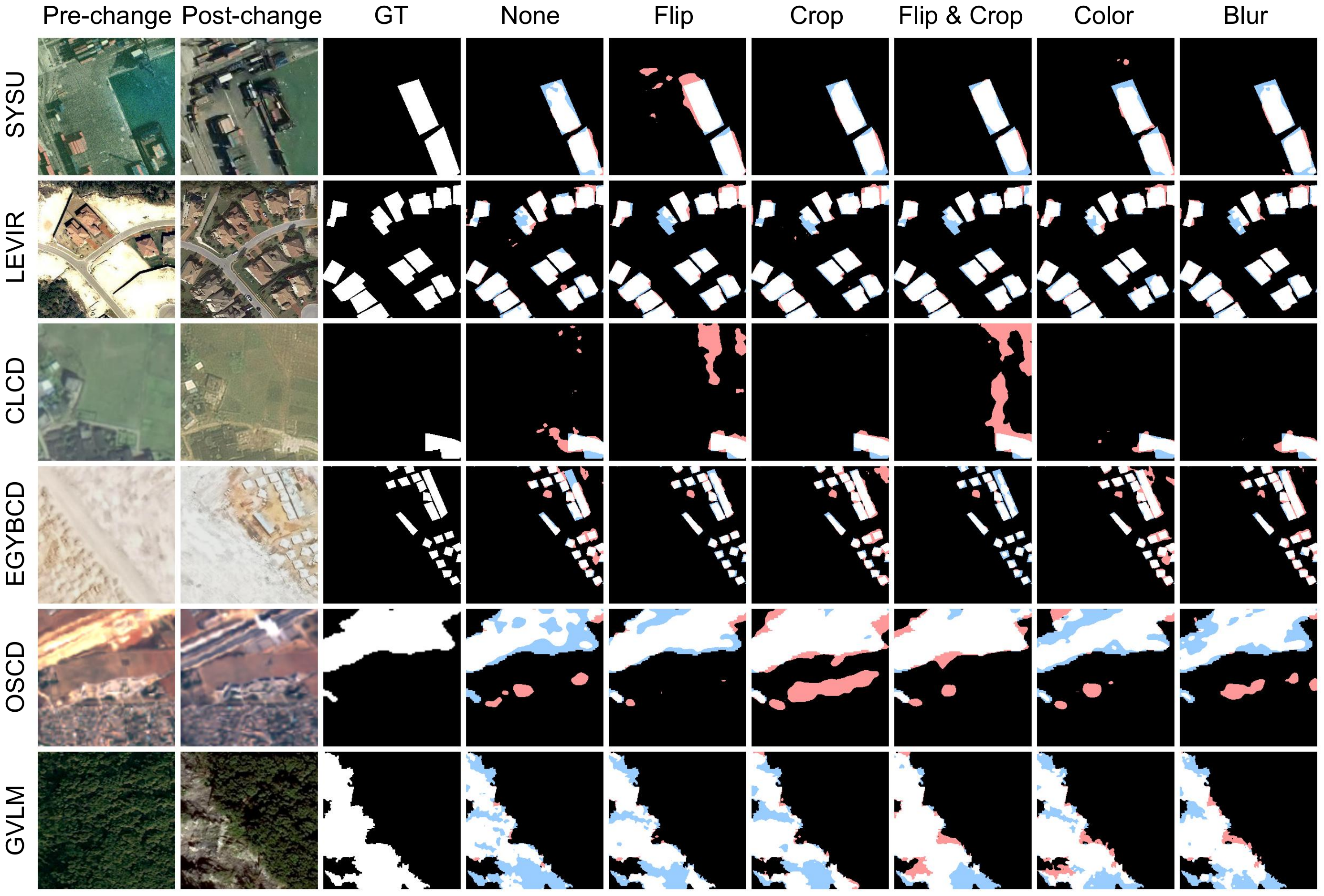}
    \caption{Additional qualitative examples for different augmentations.}
    \label{afig:qual_aug}
\end{figure*}

\begin{figure*}[!h]
    \centering
    \includegraphics[width=0.8\linewidth]{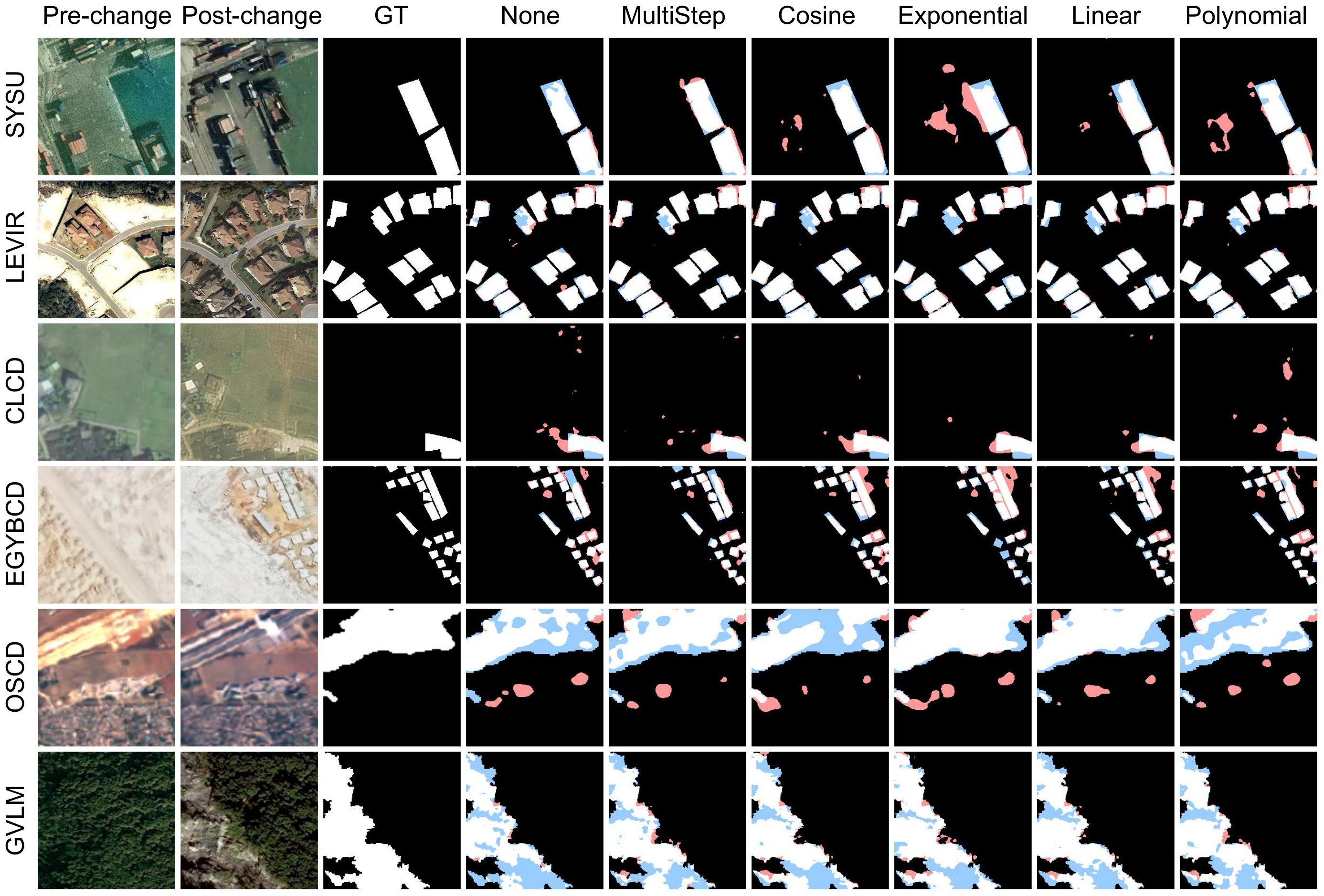}
    \caption{Additional qualitative examples for different schedulers.}
    \label{afig:qual_sc}
\end{figure*}

\begin{figure*}[!h]
    \centering
    \includegraphics[width=0.7\linewidth]{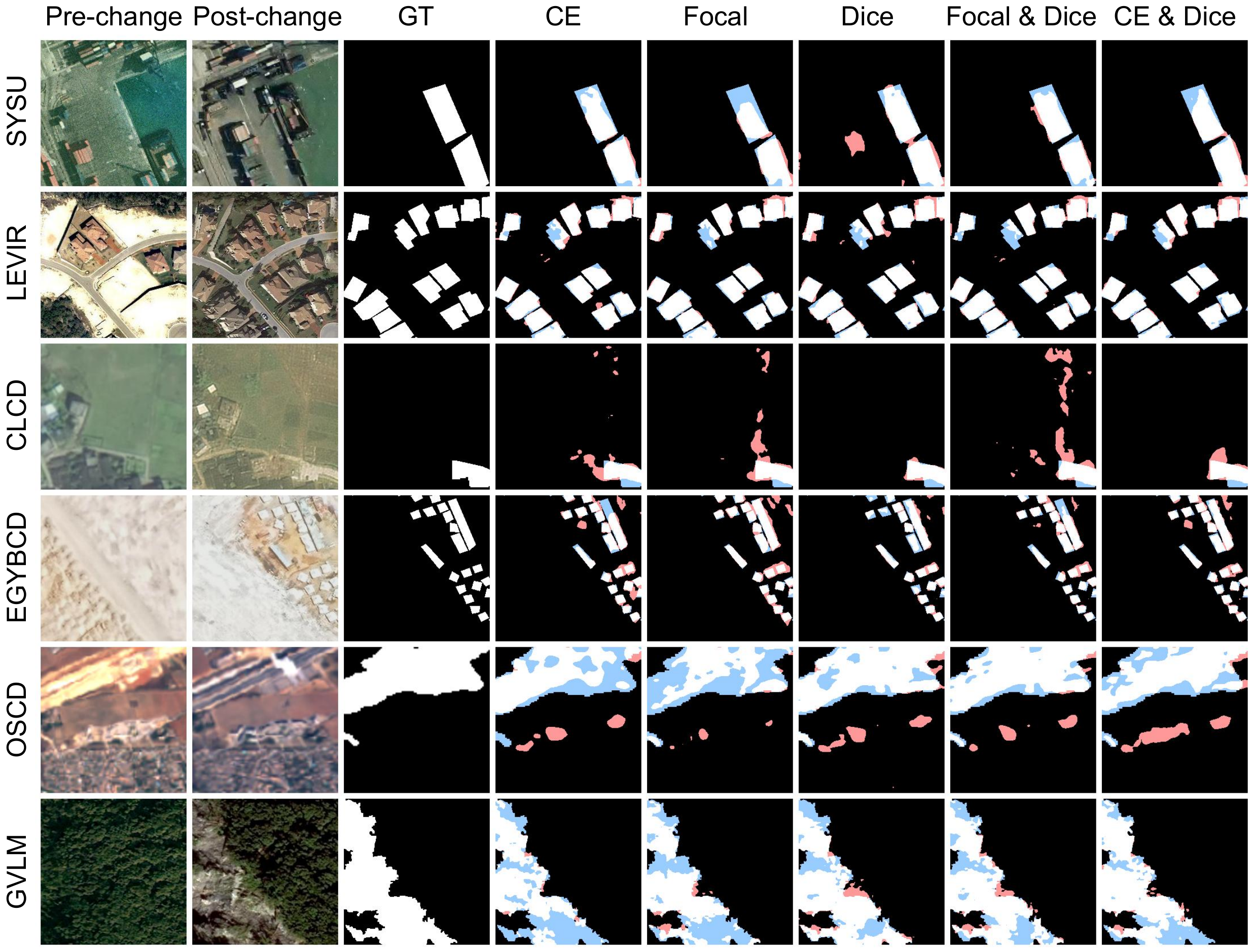}
    \caption{Additional qualitative examples for different loss functions.}
    \label{afig:qual_loss}
\end{figure*}

We also present qualitative results of all augmentations in \Cref{afig:qual_aug}, schedulers in \Cref{afig:qual_sc}, and loss functions in \Cref{afig:qual_loss}.




\subsection{State-Of-The-Art - Complete  Qualitative Results}
\label{asub:sota_qual}

\begin{figure*}[!h]
    \centering
    \includegraphics[width=0.45\linewidth]{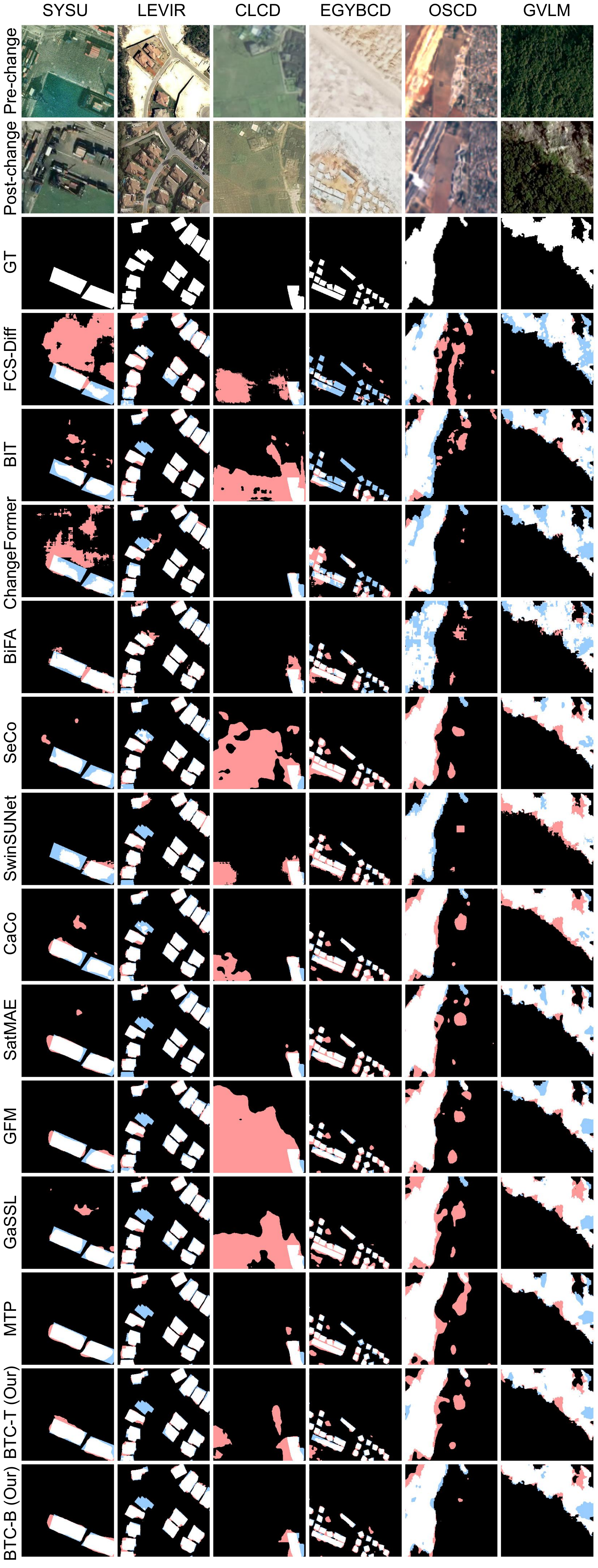}
    \caption{Additional qualitative examples for state-of-the-art methods.}
    \label{afig:qual_sota}
\end{figure*}

Qualitative results for all state-of-the-art methods are depicted in \Cref{afig:qual_sota}.

\end{document}